\documentclass[lettersize,journal]{IEEEtran}
\usepackage{amsmath,amsfonts}
\usepackage{amssymb}
\usepackage{algorithmic}
\usepackage{algorithm}
\usepackage{array}
\usepackage[caption=false,font=normalsize,labelfont=sf,textfont=sf]{subfig}
\usepackage{textcomp}
\usepackage{stfloats}
\usepackage{url}
\usepackage{verbatim}
\usepackage{graphicx}
\usepackage{cite}
\usepackage{amsmath}
\usepackage{multirow}
\usepackage{graphicx}
\usepackage{xcolor}
\usepackage{booktabs}
\usepackage{arydshln}
\usepackage{orcidlink}
\hypersetup{
    colorlinks=true,      
    linkcolor=blue,       
    citecolor=blue,       
    urlcolor=black,        
    pdftitle={Your Paper Title},
    pdfpagemode=FullScreen,
}
\begin{document}

\title{Concept Unlearning by Modeling Key Steps of Diffusion Process}
\author{Chaoshuo Zhang, Chenhao Lin, Zhengyu Zhao, Le Yang, 
Chong Zhang, Qian Wang, Chao Shen
\thanks{Chaoshuo Zhang, Chenhao Lin, Zhengyu Zhao, Le Yang, Chong Zhang, and Chao
 Shen are with the School of Cyber Science and Engineering, Xi’an Jiaotong
 University, Xi’an, China.}
\thanks{Qian Wang is with Wuhan University, Wuhan, China.}
}

\markboth{Journal of \LaTeX\ Class Files,~Vol.~14, No.~8, August~2021}%
{Shell \MakeLowercase{\textit{et al.}}: A Sample Article Using IEEEtran.cls for IEEE Journals}


\maketitle
\begin{abstract}
Text-to-image diffusion models remain susceptible to generating undesirable or harmful content. Although concept unlearning mitigates this risk, existing methods struggle with a critical optimization dilemma: thorough semantic erasure frequently induces the catastrophic forgetting of unrelated generative capabilities. To overcome this challenge, we propose Key Step Concept Unlearning (KSCU). \textcolor{black}{Serving as an integrated methodological refinement deeply motivated by information theory, KSCU explores the profound impact of step scheduling order and reveals that traditional randomized timestep sampling severely disrupts trajectory dependency.} We demonstrate that indiscriminately targeting the entire diffusion process is inefficient, as the optimal step range for unlearning inherently varies across different concepts. \textcolor{black}{Rather than globally fine-tuning all timesteps, KSCU explicitly integrates a sequential-scheduling-based Key Step Table, CFG-aware leakage compensation, and prompt augmentation to dynamically isolate optimization to a concept-specific active region.} This localized strategy successfully eradicates the target concept while preventing the structural collapse caused by early-step over-optimization. Consequently, KSCU significantly reduces computational overhead and establishes a state-of-the-art trade-off between concept erasure and utility retention. Comprehensive evaluations demonstrate that KSCU consistently delivers superior performance across diverse unlearning tasks, including nudity, style, object classes, and mass instance concepts. For example, in nudity removal, KSCU yields a 96.5\% unlearning accuracy alongside a state-of-the-art FID of 14.1. Crucially, the Key Step mechanism serves as a robust plug-and-play module. Integrating it into existing baselines inherently mitigates structural collapse while simultaneously enhancing generative preservation and computational efficiency. Code is available at  \href{https://github.com/ChaoShuoZhang/KeyStepConceptUnlearning}{KeyStepConceptUnlearning}.
\end{abstract}
\section{Introduction}
\IEEEPARstart{I}{n} recent years, the rapid advancement of text-to-image diffusion models~\cite{kingma2021variational,kawar2023imagic,dbg,ho2022classifierfree,liu2022compositional,saharia2022photorealistic} has democratized high-quality visual creation across various domains~\cite{ramesh2022hierarchical,Stability,liu2024sora}. These models empower users to translate natural language prompts into semantically coherent images. However, because large-scale diffusion models are trained on uncurated internet-scale datasets~\cite{laion,changpinyo2021conceptual,ramesh2021zero,ramesh2022hierarchical}, they inevitably memorize and regenerate problematic content, including copyrighted artworks~\cite{shan2023glaze}, culturally biased representations, and Not Safe For Work (NSFW) material~\cite{carlini2023extracting}. This unconstrained generative capability raises severe ethical, legal, and social security concerns, necessitating urgent and robust mitigation strategies.

\begin{figure}[ht]
    \centering
    \includegraphics[width=0.48\textwidth]{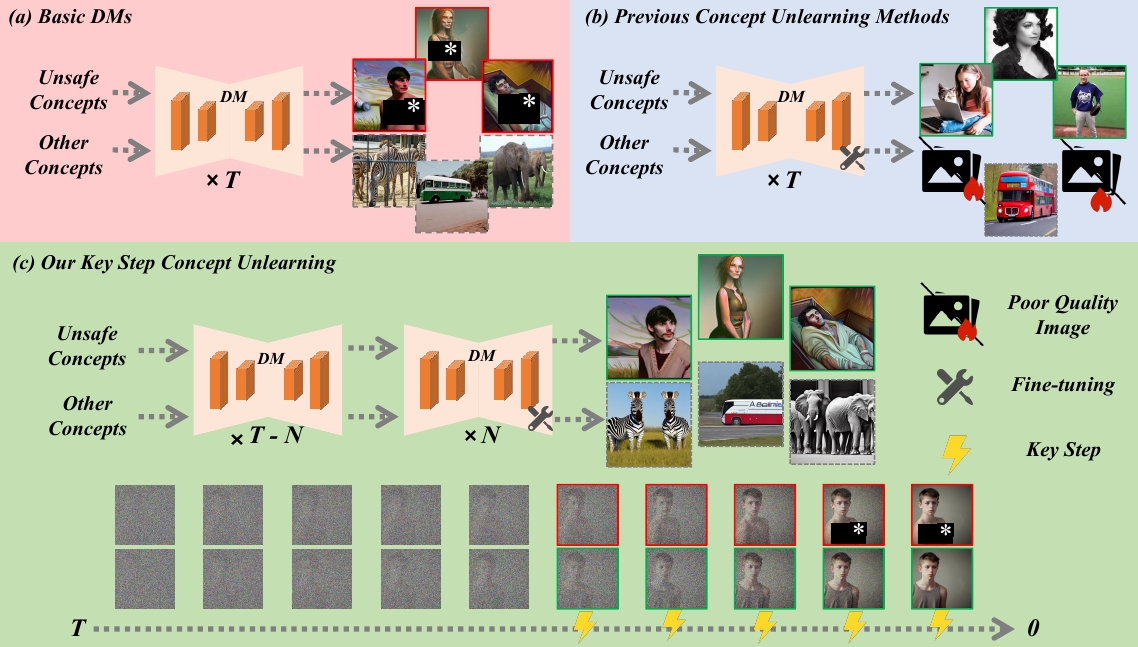}
    \caption[Illustration of Key Step Concept Unlearning]{Compared to previous methods (b) that indiscriminately optimize across the entire diffusion trajectory, KSCU (c) strictly restricts fine-tuning to an intermediate active region, achieving effective concept erasure while preventing catastrophic structural collapse.}
    \label{fig_keystep}
\end{figure}

To curb the generation of undesirable content, researchers have explored dataset screening and post-hoc filtering~\cite{rando2022redfilter,meng2022locating}. However, dataset curation is prohibitively expensive and fundamentally incomplete, while inference-time filters can be easily bypassed using adversarial prompts. Consequently, concept unlearning has emerged as a superior fine-tuning-based paradigm~\cite{ca,fan2023salun,fmn,uce,rece,wu2024erasediff}. By selectively adjusting model parameters to eradicate specific semantic mappings, diffusion models can effectively forget designated visual concepts without requiring retraining from scratch~\cite{esd,ldm}. 

Despite notable progress, existing concept unlearning methodologies suffer from two critical limitations. 
\textbf{Timestep Agnosticism.} Current methods indiscriminately fine-tune parameters across the entire reverse diffusion trajectory. However, image generation is hierarchical: early steps determine coarse global layouts, while later steps refine high-frequency semantic details~\cite{sclocchi2025phase,lee2025beta,yang2023diffusion,cao2025diffstereo}. Ignoring this temporal dynamic leads to inefficient parameter updates and destructive interference with the model's fundamental generative priors.
\textbf{The Unlearning and Retention Trade-off.} Existing methods face a severe optimization dilemma. Aggressive unlearning leads to catastrophic forgetting of unrelated concepts, whereas conservative optimization preserves overall image quality but fails to achieve thorough erasure. This trade-off remains a persistent challenge across various concept erasure tasks. State-of-the-art methods typically merely settle for a compromise between these two suboptimal extremes, rather than achieving both goals simultaneously.

\textcolor{black}{To overcome these challenges, we propose Key Step Concept Unlearning (KSCU), an integrated methodological refinement deeply motivated by information theory}. Grounded strictly in an information-theoretic analysis of frequency reconstruction dynamics, KSCU reveals that early denoising stages predominantly establish low-frequency global structures. Because high-frequency semantic details emerge exclusively later in the generative process, early-step over-optimization inherently destroys foundational structures long before the target semantics even become accessible. Furthermore, recognizing that the optimal unlearning window inherently varies across different concepts, KSCU abandons indiscriminate global parameter updates. Instead, it dynamically isolates optimization exclusively to a concept-specific active region. \textcolor{black}{Moreover, by explicitly investigating the impact of step scheduling order, we identify that randomized temporal sampling inherently disrupts the continuous sequential trajectory of diffusion models. To address this, KSCU employs a sequential-scheduling-based Key Step Table. As illustrated in Figure \ref{fig_keystep}, this dynamic algorithm enforces an incrementally advancing, forward-shifting execution order to target} these most responsive timesteps. To maximize efficacy within this restricted window, KSCU integrates a Key Step Unlearning Optimization strategy featuring a timestep-dependent dynamic weighting schedule, uniquely tailored for Classifier-Free Guidance (CFG). Finally, Prompt Augmentation is employed to enrich semantic coverage, substantially fortifying the model against adversarial prompt variations.

Our primary contributions are summarized as follows:
\begin{itemize}
\item We mathematically formalize the step-wise unlearning dynamics of diffusion models from an information-theoretic perspective. By analyzing frequency reconstruction dynamics, we rigorously demonstrate that unlearning optimization must be strictly isolated to later denoising stages. This prevents catastrophic structural collapse caused by early-step interference with low-frequency global foundations and ensures precise semantic erasure.
\item \textcolor{black}{We propose KSCU as an integrated methodological refinement deeply motivated by information theory. Uniquely, we explore the critical impact of optimization order and introduce a sequential-scheduling-based Key Step Table. This mechanism, combined with a timestep-dependent CFG-aware Unlearning Optimization schedule, and Prompt Augmentation, overcomes the trajectory gaps inherent in random sampling. This synergy enables highly localized parameter updates that constructively accumulate, with significantly reduced computational overhead.}
\item Extensive experiments across nudity, style, class, and mass concept erasure demonstrate that KSCU consistently achieves the optimal unlearning-retainment trade-off. For instance, in nudity erasure, KSCU delivers a 96.5\% unlearning accuracy alongside a state-of-the-art FID of 14.1, successfully eradicating target concepts without the severe image degradation typical of existing baselines.
\end{itemize}

\section{Background \& Related Work}
\subsection{Text-to-Image Diffusion Models}
Diffusion models~\cite{ho2020denoising} were initially proposed as a generative framework based on the reverse denoising process to reconstruct target input images. Compared with earlier paradigms such as GANs~\cite{goodfellow2020gan} and VAEs~\cite{kingma2013vae}, diffusion-based approaches demonstrate superior training stability, more faithful mode coverage, and better scalability to high-dimensional distributions. With the introduction of advanced techniques such as score-based models~\cite{song2020score}, noise-conditioned score networks~\cite{ncsm}, and denoising-based guidance~\cite{dbg}, these models have become increasingly popular for text-to-image (T2I) generation.

The emergence of Latent Diffusion Models (LDMs)~\cite{ldm} has further improved scalability and efficiency. By operating in a learned latent space, LDMs substantially reduce the computational burden while retaining high generative fidelity, making them feasible for large-scale training. Specifically, given an image $\mathbf{x}$, an encoder $E(\cdot)$ maps it to the latent code $\mathbf{z} = E(\mathbf{x})$.

The forward diffusion process then gradually corrupts the latent with Gaussian noise over $T$ steps:
\begin{equation}
\mathbf{z}_t = \sqrt{\bar{\alpha}_t} \, \mathbf{z}_0 + \sqrt{1 - \bar{\alpha}_t} \, \boldsymbol{\epsilon}, \quad \boldsymbol{\epsilon} \sim \mathcal{N}(\mathbf{0}, \mathbf{I}),
\end{equation}
where $\bar{\alpha}_t = \prod_{i=1}^{t}(1 - \beta_i)$ is the cumulative product of noise schedule coefficients.

During inference, the model reverses this process by predicting the added noise $\boldsymbol{\epsilon}$ at each step and updating the latent accordingly:
\begin{equation}
\mathbf{z}_{t-1} = \frac{1}{\sqrt{\alpha_t}} \left( \mathbf{z}_t - \frac{1 - \alpha_t}{\sqrt{1 - \bar{\alpha}_t}} \, \hat{\boldsymbol{\epsilon}}_\theta\right) + \sigma_t \, \boldsymbol{\eta}, \quad \boldsymbol{\eta} \sim \mathcal{N}(\mathbf{0}, \mathbf{I}),
\end{equation}
where $\hat{\boldsymbol{\epsilon}}_\theta$ is the predicted noise, and $\sigma_t$ is typically set according to the scheduler.

To enhance controllability, Classifier-Free Guidance (CFG)~\cite{ho2022classifierfree} interpolates between conditional and unconditional predictions:
\begin{equation}
\hat{\boldsymbol{\epsilon}}_\theta = \boldsymbol{\epsilon}_\theta(\mathbf{z}_t, t) + w \cdot \left( \boldsymbol{\epsilon}_\theta(\mathbf{z}_t, t, \mathbf{c}) - \boldsymbol{\epsilon}_\theta(\mathbf{z}_t, t) \right),
\label{eq3}
\end{equation}
where $w$ denotes the guidance scale, and $\boldsymbol{\epsilon}_\theta(\cdot)$ and $\boldsymbol{\epsilon}_\theta(\cdot, \mathbf{c})$ represent the unconditional and conditional predictions, respectively.

Together, LDMs and CFG have established diffusion models as the dominant approach in T2I generation. Their ability to generate high-quality, semantically coherent images, even under complex compositional prompts, has set a new benchmark in the field.

On this basis, recent research has focused on optimizing T2I diffusion models in terms of efficiency, scalability, and applicability. Significant breakthroughs have been made in acceleration algorithms~\cite{ddim,lu2022dpm}, model architectures~\cite{ldm}, and large-scale datasets~\cite{laion,changpinyo2021conceptual,ramesh2021zero,ramesh2022hierarchical}, all of which have contributed to the widespread adoption of these models in practical applications~\cite{ramesh2022hierarchical,Stability}. Beyond visual quality, diffusion models have also demonstrated strong generalization in multimodal and cross-domain settings, such as image editing, 3D synthesis, and video generation~\cite{blattmann2023align,poole2022dreamfusion}.

However, despite their advantages, these models are vulnerable to misuse, leading to concerns over NSFW content~\cite{rando2022redfilter}, unauthorized depictions of individuals, and copyright-infringing artwork~\cite{Stability}. Furthermore, the sheer expressive power of diffusion models introduces unique risks, such as prompt injection attacks, data leakage, and ethical dilemmas in human--AI collaboration. As T2I diffusion models continue to advance in realism and fidelity, addressing their ethical and legal challenges has become an urgent priority for the research community. Efforts in controllable generation, safety-aware fine-tuning, and concept erasure are increasingly viewed as essential to ensure their responsible deployment.

\begin{figure*}[!t]
	\centering
	\includegraphics[width=0.98\textwidth]{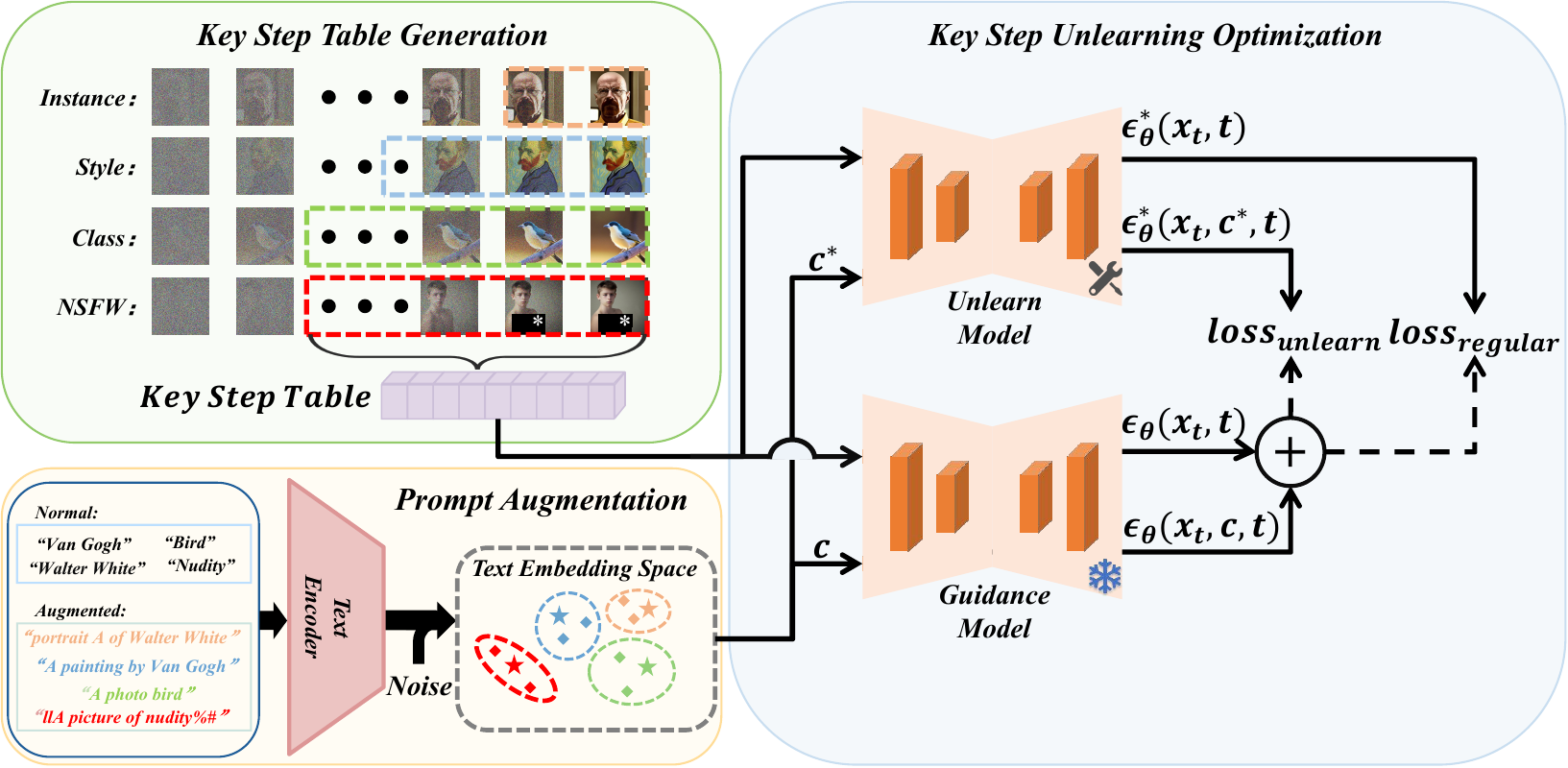}
	\caption[KSCU framework]{The KSCU framework consists of three key modules: Key Step Table, Prompt Augmentation, and Key Step Unlearning Optimization. During training, a denoising step $t$ is selected from the Key Step Table. The unlearn model performs denoising sampling to obtain $z_t$, while the Prompt Augmentation module converts the target concept $c$ into an augmented prompt $c^*$. Using $z_t$, $t$, $c$, and $c^*$, the Key Step Unlearning Optimization is computed via Equation~\eqref{eq:final_loss}, and designated parameters of the unlearn model are updated through backpropagation.}
	\label{fig:framework}
\end{figure*}

\subsection{Diffusion Model Concept Unlearning}
Large-scale datasets~\cite{laion,changpinyo2021conceptual,ramesh2021zero,ramesh2022hierarchical} often contain undesirable data, including content with inherent biases, ethically problematic information, or copyrighted materials that may later become restricted. When trained on such datasets, T2I diffusion models inevitably acquire the ability to reproduce these sensitive patterns, raising significant security and ethical concerns. To address this issue, concept unlearning has been introduced as a promising direction, aiming to selectively erase specific generative capabilities while preserving the model's ability to synthesize unrelated content—without requiring costly full retraining.

A variety of approaches have been proposed to realize this goal. Early methods such as ESD~\cite{esd} and CA~\cite{ca} rely on a frozen guiding model to provide negative supervision during fine-tuning. Building on internal representation manipulation, UCE~\cite{uce} and RECE~\cite{rece} adopt closed-form updates to cross-attention weights for efficient unlearning. Recognizing the importance of spatial and semantic localization, subsequent works heavily leverage cross-attention mechanisms~\cite{hertz2022prompt} to achieve precise concept replacement~\cite{Conceptreplacer}, localized attention-guided erasure~\cite{shilocalized}, and continuous control via LoRA sliders~\cite{Conceptsliders}. Furthermore, analyzing internal U-Net dynamics~\cite{si2024freeu} has inspired highly targeted interventions, including orthogonal value space projections~\cite{AdaVD}, high-level representation misdirection~\cite{HiRM}, and training-free gated low-rank adaptations~\cite{GLoCE}. Extensions to flow matching models via differential vector erasure~\cite{zhang2026differential} further demonstrate the versatility of training-free manipulations. To tackle the challenge of scalability, methods like SPEED~\cite{speed}, MACE~\cite{mace}, and hierarchical frameworks~\cite{tu2026mass} have significantly advanced mass concept erasure, enabling the simultaneous removal of hundreds of concepts.

Despite these methodological leaps, comprehensive benchmarking~\cite{ren2025six} reveals that existing unlearning techniques remain trapped in a severe performance dichotomy. Methods that enforce highly robust concept eradication, such as adversarial strategies~\cite{advunlearn,li2026aegis} employing challenging counterexamples, frequently over-optimize the parameter space, leading to catastrophic forgetting of unrelated generative capabilities. Conversely, techniques that strictly prioritize generative retainability often fail to achieve thorough erasure. Recent studies demonstrate that such conservative approaches leave models inherently vulnerable to adversarial prompt attacks~\cite{ringabell,p4d,unlearndiffatk} and latent space unblocking techniques capable of reawakening erased concepts~\cite{sun2026lure}. This persistent trade-off between unlearning robustness and generative retainability exposes the fragility of current frameworks, underscoring the urgent need for a more stable and balanced unlearning architecture that successfully navigates both objectives. 

While concurrent works such as ANT~\cite{ant} and SDErasure~\cite{SDErasure} also explore timestep-aware interventions for concept unlearning, KSCU distinguishes itself through a rigorous theoretical foundation. To the best of our knowledge, this is the first work to systematically analyze the critical impact of denoising steps through the lens of information theory and frequency reconstruction dynamics. Beyond empirical step-selection, our analysis elucidates the underlying mechanisms. It explicitly demonstrates how early-step over-optimization destroys low-frequency foundational structures long before high-frequency target semantics become accessible, thereby providing a definitive rationale for isolating the optimal unlearning window.
\textcolor{black}{Furthermore, prior works predominantly rely on randomized timestep sampling, a practice that severely ignores sequential trajectory dependency and introduces unpredictable trajectory gaps. To overcome this, KSCU explicitly investigates the profound impact of step scheduling order. By employing a sequential-scheduling-based Key Step Table, KSCU enforces an incrementally advancing, forward-shifting execution order. This orderly schedule ensures that parameter updates constructively accumulate, overcoming the inefficiencies of random sampling.}

\section{Key Step Concept Unlearning}
As shown in Figure~\ref{fig:framework}, our framework consists of three main components: the Key Step Table module, which controls the denoising steps during training; the Prompt Augmentation module to help the model learn a more generalized semantic distribution of the unlearned concept in the text embedding space; and the Key Step Unlearning Optimization module, including the unlearned model that undergoes fine-tuning and the guidance model that provides reverse guidance.

\subsection{Motivation \& Key Step Table Generation}

\begin{figure}[!t]
    \centering
    \includegraphics[width=0.48\textwidth]{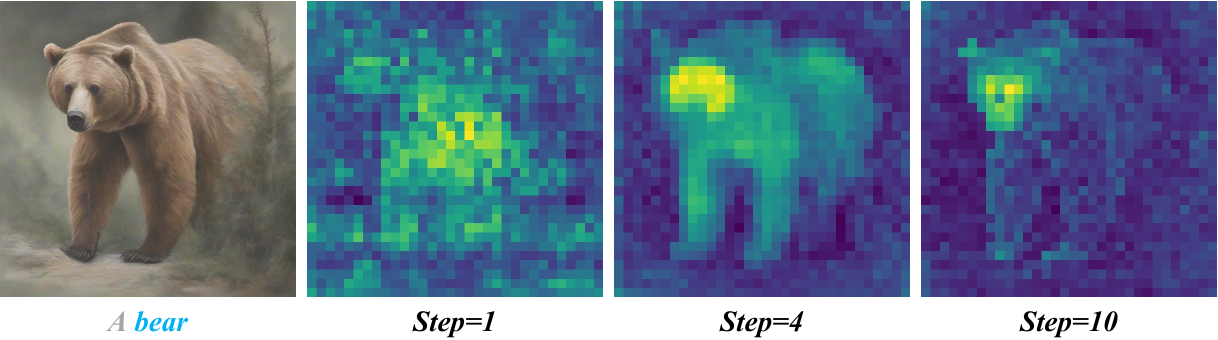}
    \caption{Correlation between text and attention map at different steps. Earlier steps determine global structures, while later steps refine localized semantic details.}
    \label{fig:attention}
\end{figure}

\begin{table}[!t]
\centering
\caption{Preliminary Verification Experiment on Stable Diffusion v1-4. The structural collapse (high FID) of ESD early 70\% steps confirms the frequency reconstruction theory. Conversely, optimizing the last 70\% steps operates within the intermediate active region, maintaining high unlearning efficacy with significantly improved generative retainability and training efficiency.}
\begin{tabular}{lcccc}
\hline
\textbf{Method} & \textbf{UA(\%)$\uparrow$}  & \textbf{UDA(\%)$\downarrow$} & \textbf{FID-30k$\downarrow$} & \textbf{Time(s)$\downarrow$} \\
\hline
ESD  & 88.2 & 73.7 & 15.4 & 2557 \\
ESD last 70\% steps   & 85.9 & 80.9 & 15.2 & 2072 \\
ESD early 70\% steps   & 90.4 & 57.7 & 69.7 & 1355 \\
\hline
\end{tabular}
\label{tab:4}
\end{table}

Due to the stepwise nature of diffusion-based generation, each sampling step contributes differently to the final image. Existing work indicates that earlier denoising steps predominantly capture low-frequency structures, while later steps progressively refine high-frequency details \cite{lee2025beta, sclocchi2025phase}. This leads to a layered emergence of semantic concepts during generation, as illustrated in Figure \ref{fig:attention}.

Beyond empirical observations, we mathematically formalize the optimal temporal window for concept erasure from an information-theoretic perspective. Viewing the reverse diffusion process through the lens of Tweedie's formula, the neural network denoiser intrinsically acts as a time-dependent Wiener filter. As rigorously proved in Appendix A and B, the emergence of spatial frequencies during generation follows a strict sequential order. Specifically, low-frequency components representing global structures are actively reconstructed strictly earlier in the reverse process. Conversely, high-frequency components containing fine-grained semantic details emerge exclusively during the later stages of denoising.

This frequency-dependent dynamic dictates a fundamental constraint for concept unlearning. At early generative stages, the high-frequency semantic details defining specific concepts are heavily suppressed by the intrinsic Wiener filter and remain mathematically inaccessible. Consequently, forcing unlearning parameter optimization during these early steps inadvertently forces the model to heavily distort the already-active low-frequency foundational bases to satisfy the erasure objective. This theoretical insight rigidly proves that unlearning interventions must be strictly isolated to the later generative timesteps to preserve global structural integrity.

To empirically validate these theoretical derivations, we conduct a preliminary verification experiment using the ESD method on Stable Diffusion v1-4. As shown in Table \ref{tab:4}, optimizing the early 70\% of steps induces severe structural collapse, yielding a degraded FID of 69.7. This confirms that early-step interference destructively alters the low-frequency foundations. Conversely, fine-tuning the last 70\% operates safely within the semantic-rich active region. This localized approach retains nearly identical unlearning accuracy to the full-step baseline while profoundly preserving generative fidelity and saving significant computational overhead.

Based on these mathematical and empirical conclusions, we design an algorithm to explicitly target this optimal active region. Recognizing that the required optimization window inherently varies across different concepts, the Key Step Table mechanism dynamically constructs an incrementally growing interval. \textcolor{black}{Crucially, this target region must be maintained as a continuous temporal interval, rather than a collection of discrete or disjoint timesteps. From an algorithmic perspective, fine-tuning isolated, discrete timesteps introduces unpredictable trajectory gaps between the modified and unmodified phases of generation. These intermittent gradient perturbations severely disrupt the continuous vector field governing the reverse diffusion process, leading to catastrophic trajectory drift and total structural collapse. Furthermore, as rigorously proven in our information-theoretic derivations (Appendices A and B), once a specific frequency emerges, its positive gradient contribution to the target concept persists continuously. Given the highly complex non-linear mappings of the network, it is technically intractable to mathematically isolate the exact, discrete semantic timesteps from this region. Therefore, maintaining a continuous intervention window is a vital algorithmic constraint to ensure stable reverse trajectory dynamics and completely eliminate the target semantics.}

This interval shifts forward to densely cover the late denoising stages. As detailed in Algorithm \ref{alg:Key Step Table}, starting from step $S$ to $E$, the full interval is appended in each iteration. After $loop_n$ iterations, the starting point $s_{\text{cur}}$ is shifted forward, and this process repeats until reaching length $L$. This specific scheduling establishes KSCU as an online learning paradigm. It ensures that the unlearning optimization densely samples from the theoretically optimal intermediate region, thereby maximizing erasure efficacy and preventing overfitting to any single discrete timestep. Crucially, we employ this incrementally advancing schedule across diffusion steps to address a fundamental trajectory dependency. Because KSCU relies on online sampling during training, optimizing the prediction $\epsilon_\theta(x_t, c, t_1)$ and subsequently altering an earlier generative step $t_2$ (where $t_2 > t_1$) fundamentally changes the reverse trajectory. This renders the previously optimized state at $t_1$ unreachable, making the prior parameter updates largely ineffective. Our incremental scheduling strictly prevents this interference.

\begin{algorithm}[!t]
\caption{Key Step Table Generation}
\label{alg:Key Step Table}
\textbf{Input}: start step $S$, end step $E$, table length $L$, full loops before shift $loop_n$\\
\textbf{Output}: $Key Step Table[:L]$
\begin{algorithmic}[1]
\STATE $Key Step Table \gets None$
\STATE $s_{\text{cur}} \gets S$
\STATE $loop_{cur} \gets 0$
\WHILE{$len(Key Step Table) < L$}
    \STATE $Key Step Table \gets Key Step Table \cup [s_\text{cur}, s_\text{cur}+1, \dots, E]$
    \STATE $loop_{cur} \gets loop_{cur} + 1$
    \IF {$loop_{cur} = loop_n$}
    \STATE $s_{\text{cur}} \gets \min(s_{\text{cur}} + 1, E - 1)$
    \STATE $loop_{cur} \gets 0$
    \ENDIF
\ENDWHILE
\end{algorithmic}
\end{algorithm}
\subsection{Prompt Augmentation}
In prompt-based concept unlearning for T2I diffusion models, the target visual concept is specified via text prompts to guide the model in identifying what to forget.  
However, accurately capturing the semantics of a complex visual concept through textual descriptions is often non-trivial. For instance, some concepts may have multiple contextual meanings, and vague or underspecified prompts may lead to incomplete unlearning, where related sub-concepts persist, or to overgeneralized forgetting, where unrelated concepts are erroneously removed.  
Recent methods~\cite{xue2025crce,chen2025safe} have recognized this limitation and introduced enhancements to the prompt design stage, often by refining textual formulations or aligning prompts with concept embeddings.  

To address this problem from a more lightweight perspective, we propose a simple yet effective Prompt Augmentation module. For a given concept $c$ to be removed, our method first leverages large language models (e.g., ChatGPT~\cite{chatgpt} and DeepSeek~\cite{deepseek}) to automatically generate a set of augmentation rules, expanding the original prompt into a diverse collection of textual variants (e.g., ``An augmented prompt about the \{target concept\}''). In practice, we typically sample ten such rules, which cover lexical, syntactic, and contextual variations of the target concept(Appendix B).  

To further enhance robustness, we incorporate additional augmentation strategies: injecting random character noise at the beginning or end of prompts (to simulate adversarial prefix/suffix attacks commonly observed in safety evaluations), shuffling word order to test semantic invariance, and removing non-essential words to reduce overfitting to specific phrasings. Beyond the textual space, we also perturb the prompt embeddings directly by adding Gaussian noise, thereby generating a dense set of representations around the target concept in the embedding space. This forces the model to learn to suppress the target concept across a continuous semantic neighborhood rather than a few discrete prompt formulations. Importantly, since the embeddings are ultimately perturbed, the strict quality of the generated textual rules is less critical—what matters is the diversity and coverage of the augmented distribution.  

Overall, Prompt Augmentation provides a lightweight yet powerful mechanism to mitigate the brittleness of text-prompt supervision in concept unlearning. By diversifying both textual and embedding-level representations, it equips the unlearning process with stronger resistance to adversarial prompts and improves generalization to real-world usage scenarios.
\subsection{Key Step Unlearning Optimization}
\begin{figure}[!t]
	\centering
	\includegraphics[width=0.48\textwidth]{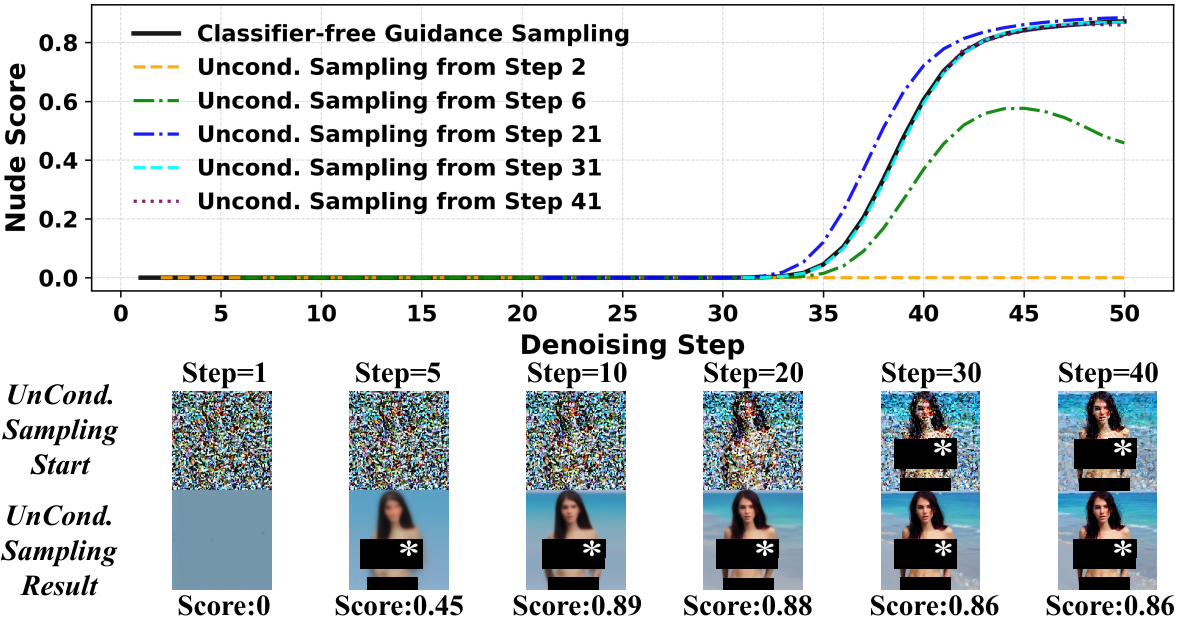}
	\caption{Image generation with Stable Diffusion v1-4 using the prompt ``naked woman": evolution of nude scores and qualitative results across the unconditional denoising process.}
	\label{fig:loss}
\end{figure}

\noindent{\textbf{Conditional Predicted Noise Optimization.}} 
According to Eq.~\eqref{eq3}, in a CFG-enabled diffusion model, the linear extrapolation between the conditional prediction $\epsilon_\theta(\mathbf{z}_t, t, \mathbf{c})$ and the unconditional prediction $\epsilon_\theta(\mathbf{z}_t, t)$ essentially approximates the gradient of the implicit classifier $\nabla_{\mathbf{z}_t} \log p(\mathbf{c} | \mathbf{z}_t)$. To effectively erase the target concept $\mathbf{c}$, our primary objective is to shift the unlearned model's conditional prediction $\epsilon^*_\theta(\mathbf{z}_t, t, \mathbf{c})$ in the exact opposite direction of this implicit classifier gradient, while preserving the generation of unrelated concepts. 

\begin{algorithm}[!t]
\caption{KSCU-Quick}
\label{alg:KSCU-quick}
\textbf{Input}: Unlearn Model $\epsilon^*_\theta$, Guidance Model $\epsilon_\theta$, scheduler $S$, target concept $c$, augmentation steps $N$, training steps $M$, Key Step Table\\
\textbf{Output}: $\epsilon^*_\theta$
\begin{algorithmic}[1]
    \STATE Sample noise $n \sim \mathcal{N}(0, 1)$
    \STATE $t \gets Key Step Table[1]$
    \STATE $z_t \gets D(\epsilon^*_\theta,n, t,0, c)$
\FOR{$i = 2$ to $M$}
    \IF{$i > N$}
        \STATE $c^* \gets \text{augment}(c)$
    \ELSE
        \STATE $c^* \gets c$
    \ENDIF
    \STATE Update parameters:\\ $\theta \gets \theta - \nabla_{\theta} \mathcal{L}_{KSCU}(\epsilon_{\theta}, \epsilon^*_{\theta}, z_t, c,c^*)$
    \IF{$i<N$}
        \STATE $t_{next}$ = $Key Step Table[i+1]$
        \IF{$t_{next}>t$}
            \STATE $z_{t_{next}}= D(\epsilon^*_\theta,z_t, t_{next},t, c)$
        \ELSE
            \STATE $z_{t_{next}}= D(\epsilon^*_\theta,n, t_{next},0, c)$
        \ENDIF
        \STATE $z_t=z_{t_{next}}$
    \ENDIF
\ENDFOR
\end{algorithmic}
\end{algorithm}

Let $\epsilon_\theta$ denote the frozen original model (acting as the guidance model) and $\epsilon^*_\theta$ denote the unlearned model undergoing fine-tuning. To achieve targeted suppression, we drive $\epsilon^*_\theta(\mathbf{z}_t, t, \mathbf{c})$ towards a negative guidance target. Setting the negative guidance scale to 1 (analogous to ESD~\cite{esd}), the conditional unlearning loss is cleanly formulated as:
\begin{equation}
\begin{split}
\text{Loss}_{\text{unlearn}} &= \left\| \epsilon_\theta^*(\mathbf{z}_t, t, \mathbf{c}) - \big( \epsilon_\theta(\mathbf{z}_t, t) - ( \epsilon_\theta(\mathbf{z}_t, t, \mathbf{c}) - \epsilon_\theta(\mathbf{z}_t, t) ) \big) \right\|_2^2 \\
&= \left\| \epsilon_\theta^*(\mathbf{z}_t, t, \mathbf{c}) - \big( 2\epsilon_\theta(\mathbf{z}_t, t) - \epsilon_\theta(\mathbf{z}_t, t, \mathbf{c}) \big) \right\|_2^2
\end{split}
\label{eq:loss_unlearn}
\end{equation}

\noindent\textbf{Unconditional Noise-Prediction Optimization.}

{\color{black}
KSCU explicitly incorporates unconditional noise prediction into the optimization objective of concept unlearning. Most existing methods formulate their objectives using condition-conditioned predictions, without explicitly constraining the model output under the null-text condition. However, although the unconditional predictor receives no text condition, its output still depends on the intermediate latent $\mathbf{z}_t$.

Let $C$ denote the concept variable. The marginal latent distribution can be expressed as a mixture of concept-conditional distributions:
\begin{equation}
p(\mathbf{z}_t)
=
\sum_{\mathbf{c}'}
p(C=\mathbf{c}')
p(\mathbf{z}_t\mid C=\mathbf{c}').
\end{equation}
Under the population-optimal solution to the standard noise-prediction objective, the unconditional prediction satisfies
\begin{equation}
\begin{split}
\epsilon_\theta(\mathbf{z}_t,t)
&\overset{\mathrm{opt}}{=}
\mathbb{E}[\epsilon\mid\mathbf{z}_t,t] \\
&=
\sum_{\mathbf{c}'}
p(C=\mathbf{c}'\mid\mathbf{z}_t,t)
\epsilon_\theta(\mathbf{z}_t,t,\mathbf{c}'),
\end{split}
\end{equation}
where
$\epsilon_\theta(\mathbf{z}_t,t,\mathbf{c}')
\overset{\mathrm{opt}}{=}
\mathbb{E}[\epsilon\mid\mathbf{z}_t,t,C=\mathbf{c}']$
denotes the corresponding concept-conditional prediction. Therefore, the contribution associated with the target concept $\mathbf{c}$ is weighted by its posterior probability $p(C=\mathbf{c}\mid\mathbf{z}_t,t)$. When preceding conditional denoising steps produce a target-bearing latent with a larger target-concept posterior, the corresponding conditional component receives a larger weight in the unconditional prediction. We refer to this propagation of target semantics through the latent-dependent unconditional prediction as Conditional Leakage.

To examine this behavior, we perform 50-step sampling using the prompt ``naked woman'' and switch from CFG to unconditional sampling at different positions $k$. As shown in Figure~\ref{fig:loss}, switching after only five conditional steps can still yield explicit nudity, while switching at $k=31$ or $k=41$ produces results visually similar to those obtained using CFG throughout. After the switch, the NudeNet score continues to increase and closely follows the CFG curve. Since all subsequent transitions are driven solely by the unconditional prediction, these results indicate that unconditional denoising continues to refine the target-related structures already encoded in $\mathbf{z}_k$, consistent with the posterior-weighted response derived above.
}

This analysis motivates explicitly constraining the unconditional prediction at target-bearing latent states. Using the Classifier-Free Guidance (CFG) heuristic, the implicit classifier gradient directing toward $\mathbf{c}$ is mathematically approximated by $\epsilon_\theta(\mathbf{z}_t, t, \mathbf{c}) - \epsilon_\theta(\mathbf{z}_t, t)$. To purify the unconditional prediction, we apply a first-order theoretical compensation. We explicitly subtract this residual conditional gradient scaled by a time-dependent coefficient $\lambda_1$:

\begin{equation}
\begin{split}
{\epsilon}_\theta^*(\mathbf{z}_t, t) &\approx \epsilon_\theta(\mathbf{z}_t, t) - \lambda_1 \big( \epsilon_\theta(\mathbf{z}_t, t, \mathbf{c}) - \epsilon_\theta(\mathbf{z}_t, t) \big) \\
&= (1+\lambda_1)\epsilon_\theta(\mathbf{z}_t, t) - \lambda_1 \epsilon_\theta(\mathbf{z}_t, t, \mathbf{c})
\end{split}
\end{equation}

We then constrain the unlearned model's unconditional branch $\epsilon^*_\theta(\mathbf{z}_t, t)$ to match this purified target via the following alignment loss:

\begin{equation}
\text{Loss}_{\text{regular}} = \left\| \epsilon_\theta^*(\mathbf{z}_t, t) - \big( (1+\lambda_1) \epsilon_\theta(\mathbf{z}_t, t) - \lambda_1 \epsilon_\theta(\mathbf{z}_t, t, \mathbf{c}) \big) \right\|_2^2
\label{eq:loss_regular}
\end{equation}

Here, $\lambda_1$ controls the compensation strength. Consistent with our frequency reconstruction analysis, as $t \to 0$, the high-frequency latent details become sharper. This causes the implicit classifier's confidence to rise drastically, exacerbating the conditional leakage into the global prior. Therefore, stronger compensation is dynamically mandated at later steps. We mathematically define $\lambda_1$ as a linear function monotonically increasing as $t$ approaches $0$. Specifically, for $t \in [1000, 0]$, we set $\lambda_1 = (1000-t) \times (2 \times 10^{-5})$.

The final Key Step Unlearning Optimization objective for KSCU is defined as:

\begin{equation}
\text{Loss} = \text{Loss}_{\text{unlearn}} + \lambda_2 \cdot \text{Loss}_{\text{regular}}
\label{eq:final_loss}
\end{equation}

where $\lambda_2 = 1 \times 10^{-4}$ provides an overall scaling factor for the leakage alignment term.

\subsection{Training Algorithms}

Although our KSCU framework significantly reduces the number of training iterations compared with conventional concept erasure approaches, the improvement in wall-clock training time remains limited. This is mainly because KSCU requires generating unlearning samples on-the-fly during fine-tuning, rather than pre-constructing the dataset as in other methods. Nevertheless, we argue that this online data generation strategy is crucial for the effectiveness of KSCU and thus remains an integral part of our design. On the other hand, we further observed that, benefiting from the step selection strategy in Algorithm~\ref{alg:Key Step Table}, the steps chosen in the KeyStepTable exhibit a periodic and incrementally increasing pattern. Inspired by this observation, we propose an acceleration strategy that leverages such regularity to skip redundant intermediate sampling steps and only perform noise propagation and state updates at critical positions. This design substantially speeds up the training process while maintaining comparable performance to the original method. Concretely, the acceleration strategy reuses forward trajectories and recomputes states only at periodic checkpoints, effectively eliminating the computational overhead of repeated step-wise sampling. The complete KSCU algorithm, including this acceleration strategy, is summarized in Algorithm~\ref{alg:KSCU-quick}.

\section{Experimental Results}

\subsection{Experimental Setup}
\noindent\textbf{Dataset.}  
We evaluate concept unlearning across five distinct scenarios: class, style, nudity, single instance, and mass concept erasure. We utilize two primary datasets, I2P~\cite{i2p} and UnlearnCanvas~\cite{zhang2024unlearncanvas}, alongside a curated celebrity dataset.

For nudity erasure, we sample a subset of I2P containing 931 prompts labeled as sexual. Using NudeNet~\cite{nudenet} with a 0.2 threshold, this subset yields 575 detected nudity regions, effectively capturing the original semantic distribution while optimizing computational efficiency. 

UnlearnCanvas is a high-resolution stylized image dataset with 20 classes and 60 styles. \textcolor{black}{To balance computational constraints with evaluation depth, we utilized a highly representative subset of 10 diverse classes and 10 representative styles:}

\begin{itemize}  
    \item \textbf{Classes}: Architectures, Bears, Cats, Dogs, Humans, Towers, Flowers, Sandwiches, Trees, Sea.  
    \item \textbf{Styles}: Abstractionism, Artist Sketch, Cartoon, Impressionism, Monet, Pastel, Pencil Drawing, Picasso, Sketch, Van Gogh.  
\end{itemize}  

\textcolor{black}{This subset was strategically composed to maximize semantic diversity: the chosen classes cover animate entities, inanimate natural objects, and man-made structures, while the styles span distinct master painters (e.g., ``Monet''), broad art movements, and illustrative line arts (e.g., ``Sketch''). This orthogonal design ensures that our unlearning framework is rigorously stress-tested across features of vastly different structural complexities.}

For instance and mass concept unlearning, we utilize the established celebrity dataset introduced by MACE~\cite{mace}, comprising up to 200 specific identities. This allows us to rigorously stress-test the model's scalability, ranging from single-instance erasure to extreme multi-concept interference.
\begin{figure*}[!t]
    \centering
    \includegraphics[width=0.24\textwidth]{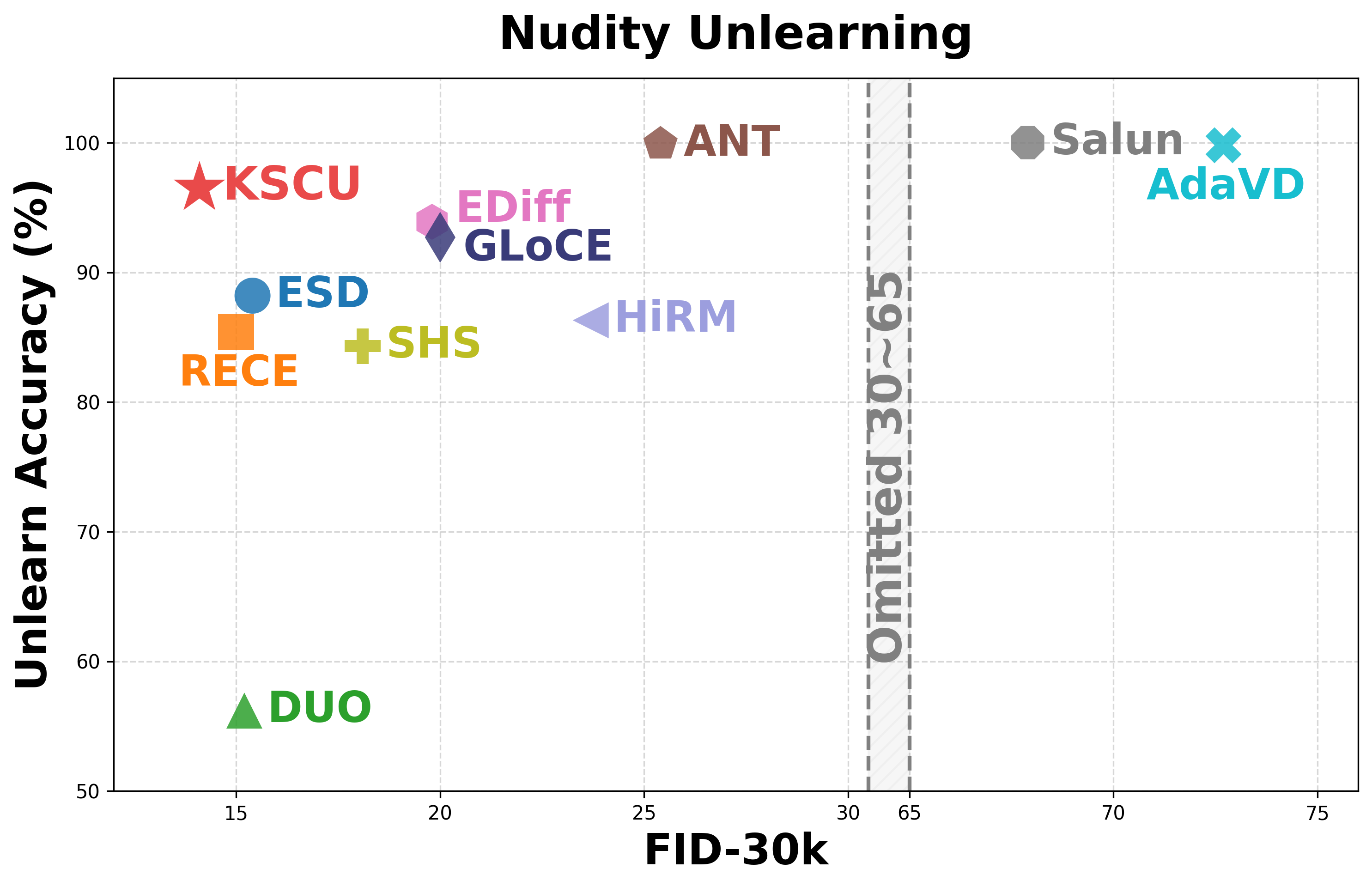}
    \includegraphics[width=0.24\textwidth]{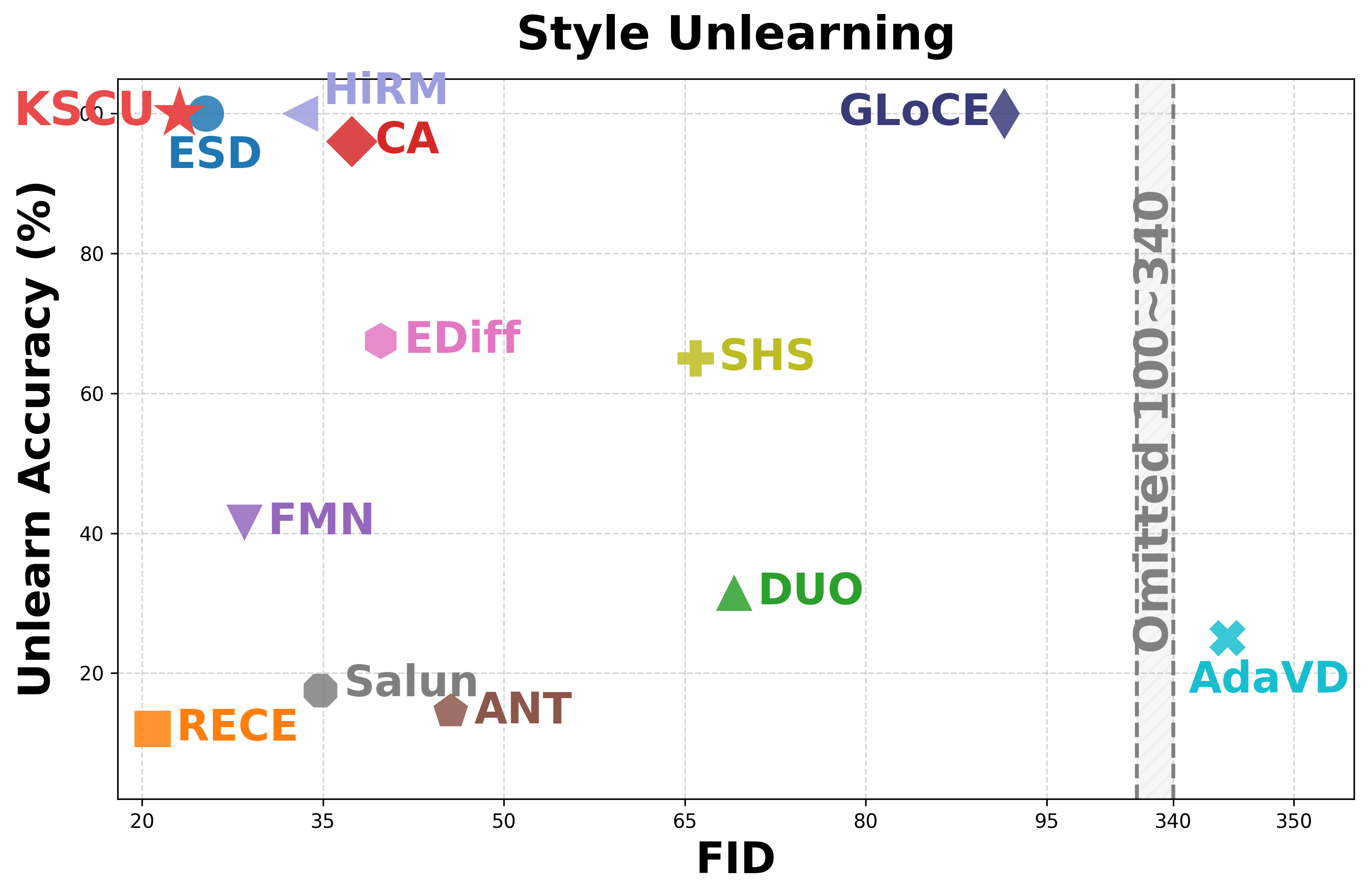}
    \includegraphics[width=0.24\textwidth]{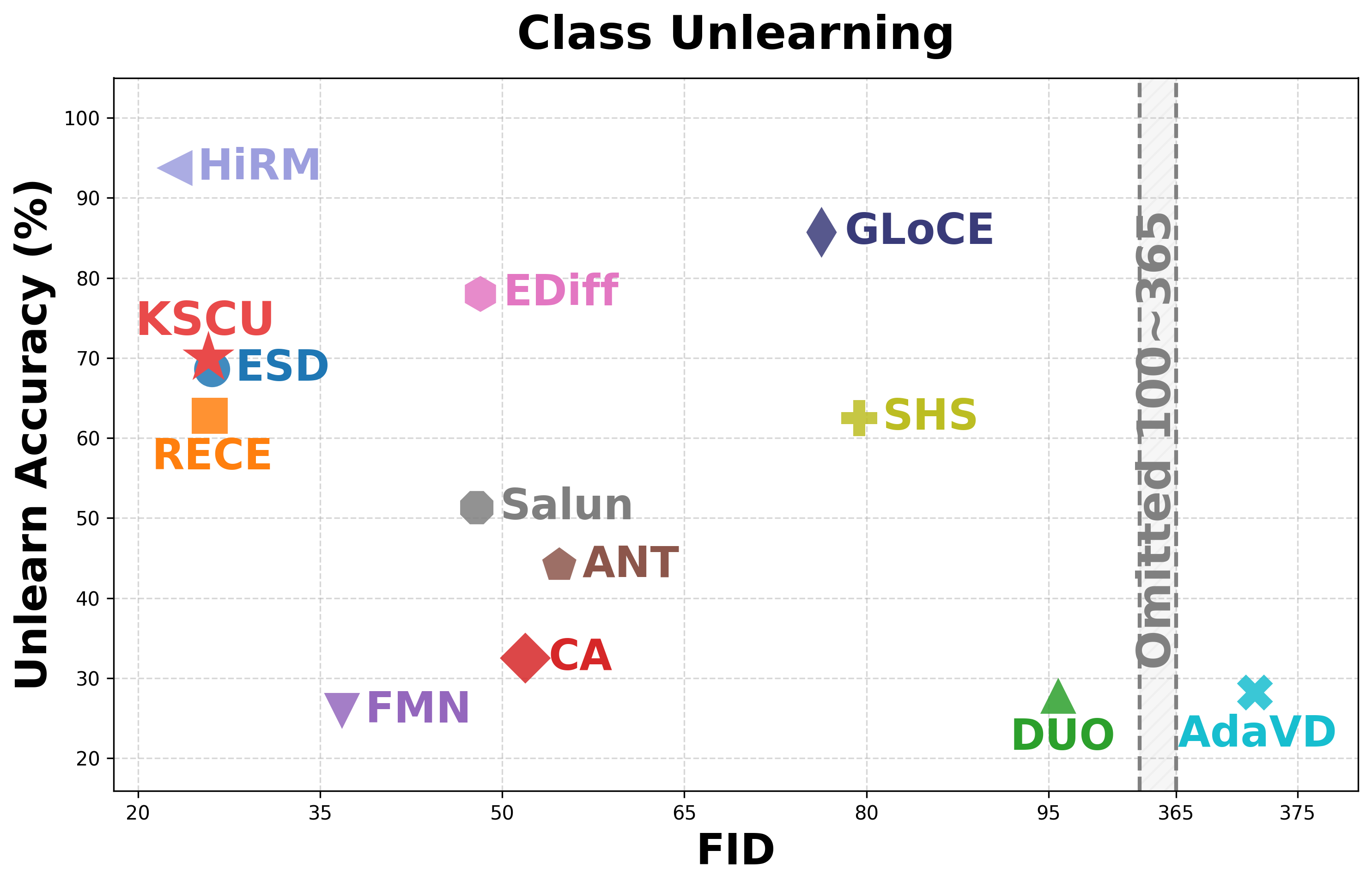}
    \includegraphics[width=0.24\textwidth]{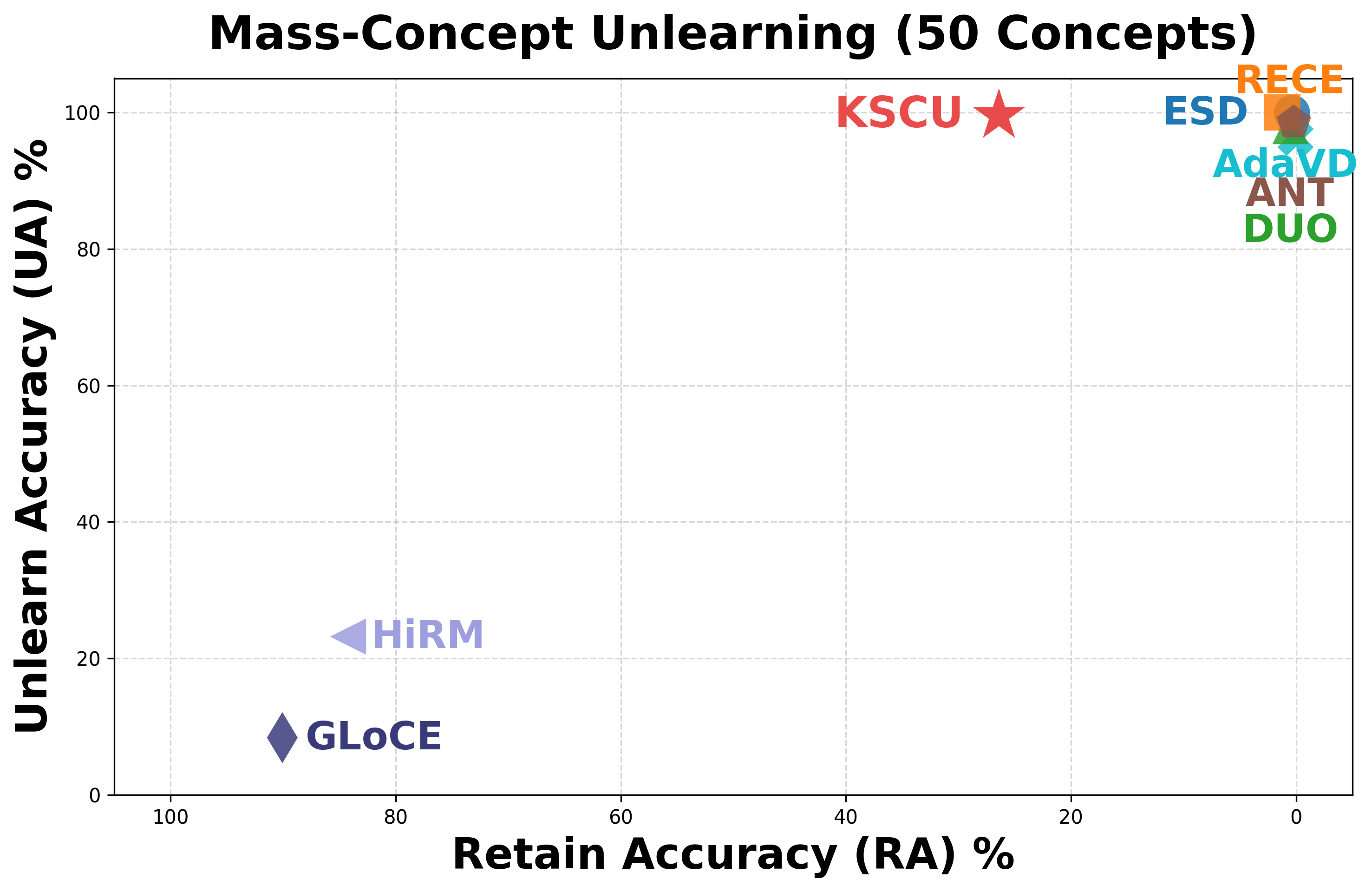}
	\caption{Performance trade-off landscapes across four unlearning tasks: nudity, style, class, and mass concept (50 concepts) erasure. Models positioned closer to the top-left region exhibit a superior balance between unlearning efficacy and generative retainability.}
	\label{fig:overview}
\end{figure*}

\begin{figure*}[!t]
	\centering
	\includegraphics[width=0.98\textwidth]{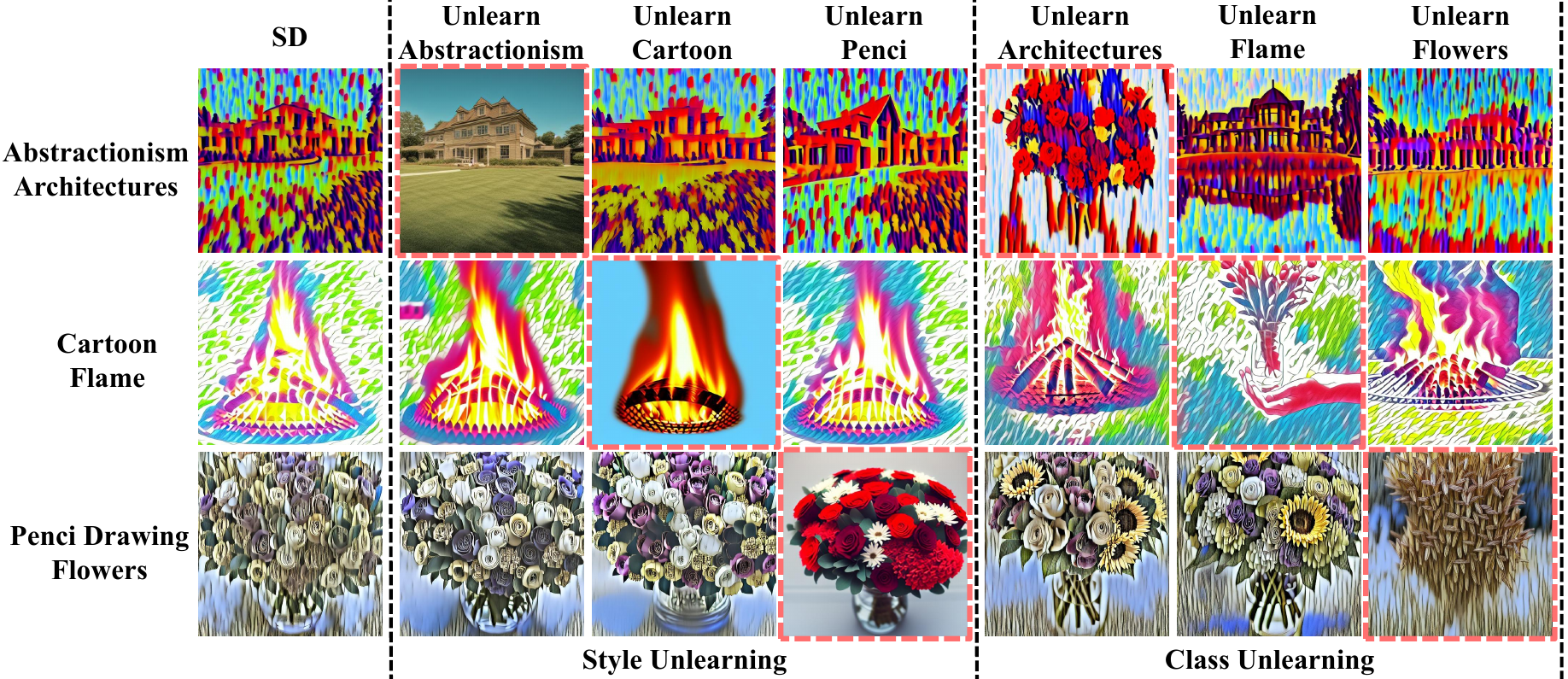}
	\caption{KSCU's Qualitative Performance on {UnlearnCanvas}. Red-bordered images show the generations of the unlearned concept after KSCU, while borderless images depict the results of unrelated concepts.}
	\label{fig:style_artists}
\end{figure*}

\noindent\textbf{Evaluation.}  
We adopt Unlearn Accuracy (UA)~\cite{esd} as the evaluation metric, computed as $\text{UA} = 1 - x/y$. In {I2P}, $x$ is the number of nudity regions detected post-unlearning, and $y = 575$. Unlike prior methods~\cite{esd} that report the number of nude parts, we aggregate all detections into a single UA score for clarity and comparability. In {UnlearnCanvas}, $x$ is the number of classifier-detected target concepts, and $y$ is the number of generated images for the corresponding prompts.

We also report Fréchet Inception Distance (FID)~\cite{fid} to measure image quality. For {I2P}, prompts are drawn from {COCO}~\cite{coco}, and FID is computed against real images from coco30k. For {UnlearnCanvas}, images are generated with the prompt ``A \{class\} image in \{theme\} style," and compared to real images in the dataset. 
For {UnlearnCanvas}, we further report In-domain Retain Accuracy (IRA) and Cross-domain Retain Accuracy (CRA) to assess the preservation of non-target information. To evaluate defense performance, we generate adversarial prompts using UDA~\cite{unlearndiffatk}, reporting Attack Success Rate (ASR) on {I2P}.

Specifically, for mass concept unlearning, we utilize Unlearn Accuracy (UA) and Retain Accuracy (RA). UA quantifies the successful erasure of the targeted celebrity identities (i.e., recognition failure), while RA measures the preservation of identity fidelity and generative quality for non-target concepts. 

Rather than relying on a heuristic aggregated score, we rigorously evaluate the overall performance by analyzing the trade-off landscape between UA and generative retainability (e.g., FID or RA). This comparative scatter-plot approach provides a direct, transparent, and unbiased assessment of the unlearning-retainment trade-off without the bias introduced by metric engineering.

\noindent\textbf{Baselines and Training Details.}  
To ensure a comprehensive and rigorous evaluation across diverse unlearning scenarios, we benchmark KSCU against an extensive suite of 12 recent state-of-the-art methods: ESD~\cite{esd}, RECE~\cite{rece}, CA~\cite{ca}, FMN~\cite{fmn}, ANT~\cite{ant}, DUO~\cite{park2024direct}, EDiff~\cite{wu2024erasediff}, Salun~\cite{fan2023salun}, SHS~\cite{shs}, AdaVD~\cite{AdaVD}, GLoCE~\cite{GLoCE}, and HiRM~\cite{HiRM}. All baselines utilize the optimal hyperparameters reported in their respective original papers.

For our proposed KSCU, training is conducted with a batch size of 1 over 50 denoising timesteps. During style unlearning, the Key Step Table initializes at starting step $S=25$ with $loop_n=8$, selectively updating only the cross-attention modules for 500 optimization iterations. For all other tasks, the table initializes at $S=15$ with $loop_n=8$, updating all U-Net modules except cross-attention for 700 iterations. We utilize the Adam optimizer with a learning rate of $1 \times 10^{-5}$. DDIM sampling with 50 steps is employed across all experiments. Stable Diffusion v1.5 is adopted for UnlearnCanvas, while all other tasks utilize v1.4 unless otherwise specified. All experiments are executed on a single NVIDIA L40 GPU.

\subsection{Tradeoff Overview}

As illustrated in Figure \ref{fig:overview}, we map the performance trade-off landscapes across four distinct unlearning scenarios. The scatter plots clearly demonstrate that KSCU establishes the optimal trade-off in both nudity and style unlearning tasks, consistently anchoring itself in the ideal top-left region. For class unlearning, our method remains highly competitive, delivering a trade-off capacity that closely rivals HiRM. Furthermore, in the extremely challenging mass concept unlearning setting, KSCU achieves superior generative retainability among all competitors capable of maintaining robust unlearning efficacy.

\subsection{Class and Style Unlearning}

\begin{table*}[ht]
\centering
\caption{Class and Style Unlearning Quantitative Performance on UnlearnCanvas. The bold font represents the best, and the underlined font represents the second-best.}
\begin{tabular}{lcccc|cccc}
\hline
\multirow{3}{*}{\textbf{Method}} & \multicolumn{4}{c|}{\textbf{Class}} & \multicolumn{4}{c}{\textbf{Style}} \\
\cline{2-9}
& \multicolumn{1}{c|}{\textbf{Effective}} & \multicolumn{3}{c|}{\textbf{Generative Retainability}} & \multicolumn{1}{c|}{\textbf{Effective}} & \multicolumn{3}{c}{\textbf{Generative Retainability}} \\
\cline{2-9}
& \multicolumn{1}{c|}{\textbf{UA(\%)$\uparrow$}} 
& \multicolumn{1}{c|}{\textbf{IRA(\%)$\uparrow$}}
& \multicolumn{1}{c|}{\textbf{CRA(\%)$\uparrow$}}
& \multicolumn{1}{c|}{\textbf{FID$\downarrow$}} 
& \multicolumn{1}{c|}{\textbf{UA(\%)$\uparrow$}} 
& \multicolumn{1}{c|}{\textbf{IRA(\%)$\uparrow$}}
& \multicolumn{1}{c|}{\textbf{CRA(\%)$\uparrow$}}
& \multicolumn{1}{c}{\textbf{FID$\downarrow$}} \\
\hline
ESD~\cite{esd}(ICCV'23)      & 68.6 & \underline{98.9} & \underline{96.4} & 26.1 & \textbf{100.0} & 85.7 & \underline{99.0} & 25.3 \\
RECE~\cite{rece}(ECCV'24)      & 62.8 & 88.9 & 96.3 & {25.9} & 12.0 & \textbf{97.6} & 98.4 & \textbf{20.9} \\
DUO~\cite{park2024direct}(NeurIPS'24)   & 27.8 & 89.9 & 75.2 & 95.8 & 31.5 & 77.0 & 89.8 & 69.1 \\
CA~\cite{ca}(ICCV'23)         & 32.5 & 86.7 & 83.4 & 51.9 & 96.0 & 86.6 & 88.7 & 37.4 \\
FMN~\cite{fmn}(CVPR'24)      & 25.9 & 93.0 & 96.1 & 36.8 & 41.5 & 95.1 & 91.3 & 28.5 \\
ANT~\cite{ant}(MM'25)      & 44.1 & 80.7 & 89.6 & 54.7 & 14.5 & 91.7 & 78.2 & 45.6 \\
EDiff~\cite{wu2024erasediff}(CVPR'25) & 78.0 & 94.1 & 90.1 & 48.2 & 67.5 & 93.7 & 98.3 & 39.8 \\
Salun~\cite{fan2023salun}(ICLR'24) & 51.3 & 93.5 & 94.6 & 47.9 & 17.5 & \underline{96.3} & 95.2 & 34.8 \\
SHS~\cite{shs}(ECCV'24) & 62.5 & 54.6 & 56.7 & 79.4 & 65.0 & 51.1 & 29.4 & 65.9 \\
AdaVD~\cite{AdaVD}(CVPR'25) & 28.2 & 4.7  & 1.9  & 371.5& 25.0 & 1.8  & 5.0  & 344.5\\
GLoCE~\cite{GLoCE}(CVPR'25) & \underline{85.7} & 65.2 & 55.9 & 76.3 & \textbf{100.0}& 49.4 & 76.4 & 91.5 \\
HiRM~\cite{HiRM}(ICLR'26)                & \textbf{93.7} & 96.7 & 79.7 & \textbf{23.0} & \textbf{100.0}& 81.2 & 97.3 & 33.1 \\
KSCU                & {70.0} & \textbf{99.0} & \textbf{97.1} & \underline{25.8} & \textbf{100.0} & 88.7 & \textbf{99.3} & \underline{23.1} \\
\hline
\end{tabular}
\label{tab:comparison_methods}
\end{table*}

To evaluate the effectiveness of our proposed KSCU, we first perform qualitative assessments on class and style unlearning tasks. As illustrated in Figure \ref{fig:style_artists}, KSCU successfully blocks the generation of target concepts (red-bordered images) while exhibiting minimal disruption to unrelated classes and styles (borderless images). The quantitative evaluation is detailed below.

\textbf{Class Unlearning. }
Evaluating a diffusion unlearning method requires a holistic analysis of the trade-off between conceptual erasure (UA) and generative retainability (IRA, CRA, and FID). As shown in Table \ref{tab:comparison_methods}, while recent baselines like HiRM and GLoCE achieve high UA scores, they suffer from severe degradation in cross-domain retainability, with CRA dropping to 79.7\% and 55.9\%, respectively. Highly aggressive erasure often collapses the model's generative diversity. Conversely, KSCU establishes an exceptional trade-off. It achieves a highly competitive UA of 70.0\% while setting the state-of-the-art in generative preservation, yielding the highest IRA (99.0\%) and CRA (97.1\%). Furthermore, KSCU maintains a strong FID of 25.8, decisively outperforming high-UA methods like EDiff (FID 48.2) and GLoCE (FID 76.3). This confirms that KSCU effectively suppresses target classes without catastrophically compromising image quality and diversity.

\textbf{Style Unlearning.} 
As detailed in Table \ref{tab:comparison_methods}, KSCU demonstrates remarkable efficacy in style unlearning. KSCU achieves perfect erasure (UA of 100.0\%) alongside ESD, GLoCE, and HiRM. Crucially, among these highly effective methods, KSCU significantly outperforms its competitors in generative retainability. It achieves the highest CRA (99.3\%) and an outstanding FID of 23.1. While RECE reports a marginally better FID (20.9), it fundamentally fails at the primary unlearning task (UA of 12.0\%), securing image quality solely by avoiding concept forgetting. 
These results emphasize the architectural advantage of our approach. By selectively fine-tuning only the later stages of the denoising trajectory, KSCU reduces the required training iterations by half compared to full-step methods like ESD. This localized optimization strictly limits the disruption to the model's inherent generative priors, offering a more efficient, controllable, and scalable solution for concept unlearning in large text-to-image diffusion models.

\subsection{Nudity Unlearning \& Robustness}
\begin{table}[!t]
\centering
\caption{Evaluation of Unlearning Capability on the Nudity task. Time refers to the execution duration of a single unlearning run, with certain baselines incorporating the time required for data generation.}
\begin{tabular}{lcccc}
\hline
\textbf{\multirow{2}{*}{Method}} & \multicolumn{2}{c}{\textbf{Effective}} & \multicolumn{1}{c}{\textbf{Retain}} & \multicolumn{1}{c}{\textbf{Speed}} \\
\cline{2-5}
  &\multicolumn{1}{c}{\textbf{UA(\%)$\uparrow$}}  &\multicolumn{1}{c}{\textbf{UDA(\%)$\downarrow$}} & \multicolumn{1}{c}{\textbf{FID-30k$\downarrow$}} &\multicolumn{1}{c}{\textbf{Time(s)$\downarrow$}} \\
\hline
ESD      & 88.2 & 73.7 &15.4 &2557\\
RECE     & 85.4 & 87.5 &\underline{15.0} &{22}\\
Salun    & \textbf{100.0} & \textbf{4.2} &67.9 &581\\
DUO      & 56.2 & 98.3 &15.2 &628\\
ANT      & \underline{99.9}& \underline{5.3}&25.4&3016\\ 
EDiff    & 93.9 & 52.5 &19.8 &1914\\
SHS      & 84.3 & 95.5 &18.1 &817\\
AdaVD    & 99.8 & 9.7 &72.7 & \underline{10}\\
GLoCE    & 92.7 & 62.2    &20.0 &\textbf{6}\\
HiRM     & 86.3 & 61.5 &23.7 &17\\
KSCU     & 96.5 & 47.2 &\textbf{14.1} &{546}\\
\hline
\end{tabular}
\label{tab:2}
\vspace{-0.5em}
\end{table}
\begin{figure*}[!t]
	\centering
    \includegraphics[width=0.9\textwidth]{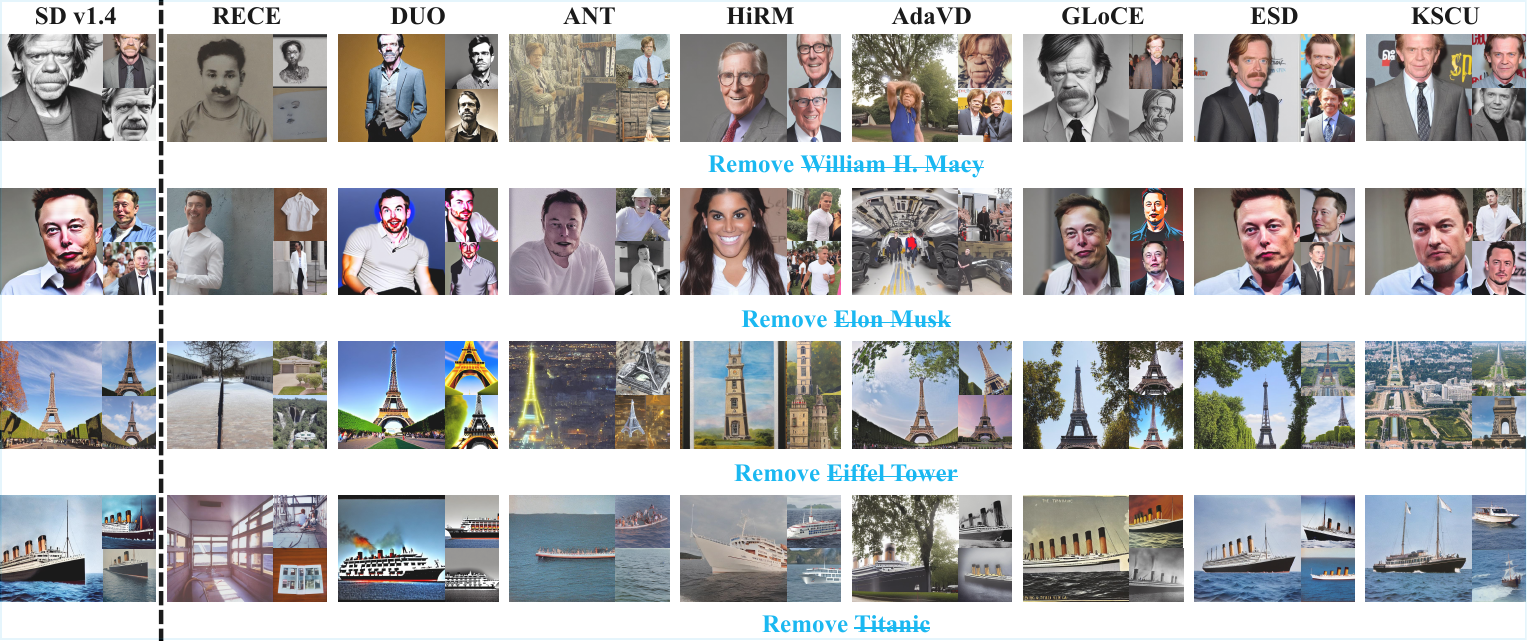}
	\caption{Qualitative comparison of instance unlearning across four distinct concepts. KSCU consistently achieves more thorough target erasure while strictly preserving the generative quality and semantic integrity of unrelated visual elements.}
	\label{fig:instance}
\end{figure*}

\noindent{\textbf{Nudity Unlearning.}}  
In the NSFW concept unlearning experiment, we fine-tune models to completely erase the concept of nudity. The evaluation is conducted by generating 931 images using prompts from the i2p~\cite{i2p} dataset and 30,000 images using COCO~\cite{coco} prompts. We measure unlearning efficacy using Unlearn Accuracy (UA) alongside an adversarial Attack Success Rate (UDA), while generative retainability is assessed via FID-30K. 

As detailed in Table \ref{tab:2}, KSCU achieves a superior balance between unlearning effectiveness and generative retainability. Specifically, KSCU attains a UA of 96.5\%, aggressively suppressing NSFW content, while simultaneously achieving an FID-30K of 14.1—the lowest among all evaluated methods. In contrast, while Salun, ANT, and AdaVD achieve near-perfect UA scores (over 99\%), their FID scores severely degrade (e.g., 67.9 for Salun and 72.7 for AdaVD). Visual inspections reveal extensive artifacts and large-area distortions in their generations, indicating catastrophic over-unlearning. 
Conversely, methods that manage to maintain competitive FID scores, such as RECE (15.0) and DUO (15.2), exhibit insufficient unlearning efficacy, with UA dropping to 85.4\% and 56.2\%, respectively. By navigating this trade-off, KSCU establishes itself as the most effective and balanced approach for nudity unlearning. Furthermore, while closed-form or lightweight methods like RECE and GLoCE execute faster, they significantly lag behind KSCU in overall trade-off capability. Compared to standard fine-tuning approaches like ESD (2557s) and ANT (3016s), KSCU dramatically reduces the required training time to 546s, offering a highly practical unlearning solution.

\noindent{\textbf{Robustness.}}  
To evaluate adversarial defense capabilities, we utilize UDA~\cite{unlearndiffatk} to generate adversarial continuous prompts designed to bypass the unlearning mechanisms. When examining the attack success rate (lower UDA is better), we exclude Salun and ANT from the robustness comparison, as their exceptionally low UDA scores (4.2\% and 5.3\%) are merely artifacts of severe over-unlearning and model collapse, rather than true defensive robustness. Among all functionally viable SOTA methods, KSCU achieves the lowest UDA score of 47.2\%. 
Furthermore, we observe a distinct correlation between a method's baseline unlearning precision and its resilience to adversarial attacks. We attribute this phenomenon to the curated nature of the i2p dataset, which comprises diverse prompts with inherent adversarial characteristics. Because thorough and precise erasure fundamentally removes the latent semantic hooks that adversarial algorithms attempt to exploit, strong unlearning performance on such challenging datasets serves as a reliable proxy for defensive capability. Consequently, KSCU's superior unlearning efficacy naturally provides a more robust shield against complex, real-world adversarial manipulations.

\subsection{Instance and Mass Concept Unlearning}
\begin{figure}[!t]
    \centering
    \includegraphics[width=0.24\textwidth]{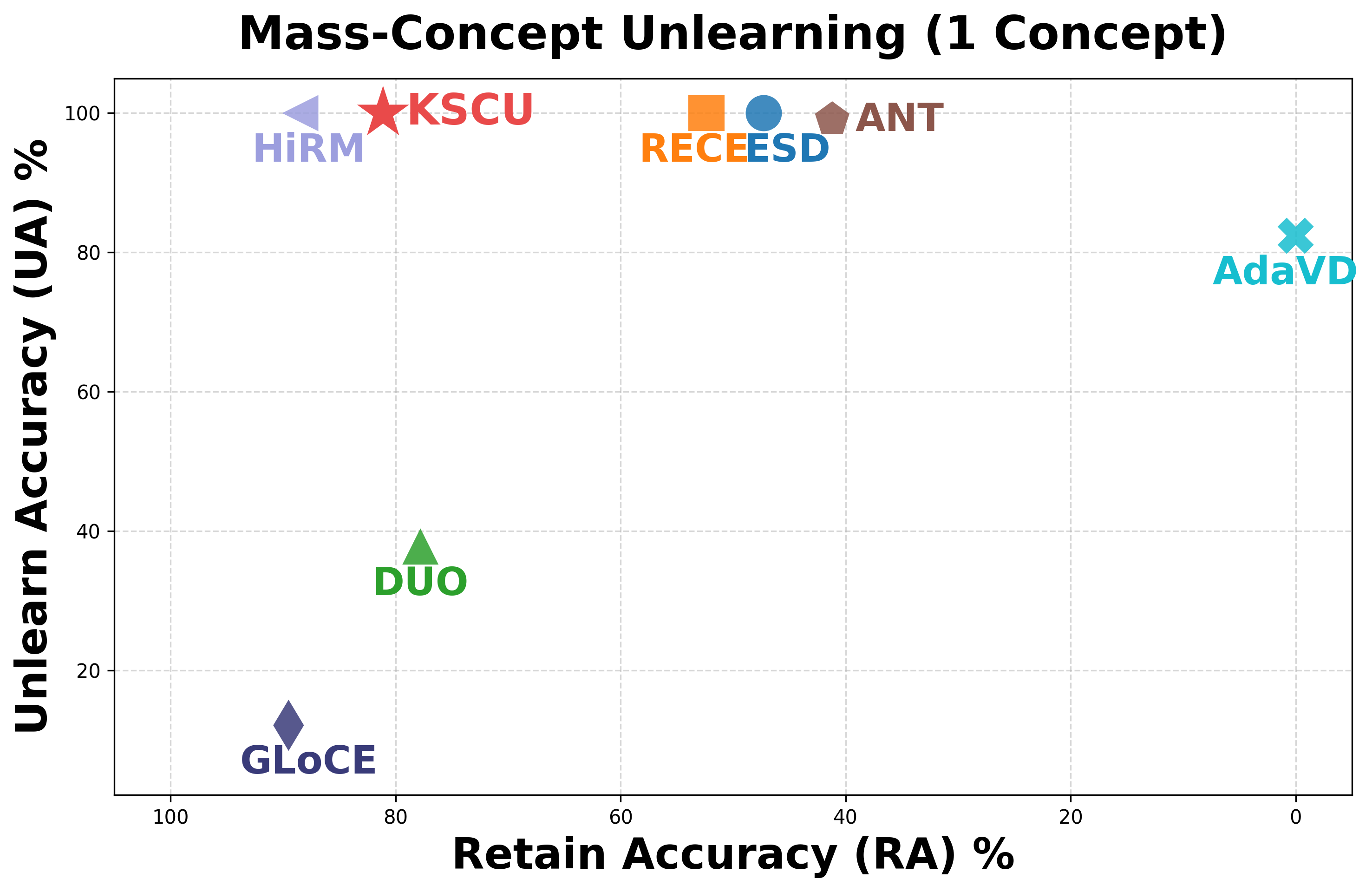}
    \includegraphics[width=0.24\textwidth]{result/tradeoff_ua_ra_50c_inverted.png}
    \caption{Performance trade-off landscapes for Mass Concept Unlearning, comparing a single target to an extreme 50-concept load. The x-axis (Retain Accuracy) is inverted (Lower is worse) to align with the y-axis (Unlearn Accuracy), meaning models positioned in the top-left region achieve the optimal balance. KSCU demonstrates superior scalability and stability under the extreme 50-concept setting.}
    \label{fig:mass}
\end{figure}
\textbf{Single Instance Unlearning.} 
To further demonstrate KSCU's efficacy in erasing specific memorized instances, we conduct qualitative comparisons against state-of-the-art approaches. As illustrated in Figure \ref{fig:instance}, KSCU not only successfully eradicates target instances (e.g., public figures and landmarks) but also exhibits a unique semantic substitution mechanism. Instead of indiscriminately corrupting the target region or leaving unnatural voids, KSCU replaces the erased instance with visually plausible and contextually coherent alternatives (e.g., substituting the ``Eiffel Tower'' with a structural ``highway'' or the ``Titanic'' with a generic ``small boat''). 
In contrast, while methods like HiRM and RECE forcefully suppress target identities, they introduce severe distributional shifts and background distortions. Conversely, DUO and GLoCE fail to achieve sufficient erasure, leaving recognizable traces of the original instances. By achieving highly localized forgetting with only 200 parameter updates, KSCU significantly reduces computational overhead, demonstrating strong potential for scalability.

\noindent\textbf{Mass Instance Concept Unlearning.} 
Building upon the superior performance of KSCU on single instance concept tasks, we introduce mass instance concept experiments to rigorously stress-test the model against up to 50 instance concepts. As illustrated in the scatter plot of Figure \ref{fig:mass}, KSCU and baselines like HiRM initially cluster in the optimal top-left region during single-concept erasure. However, scaling to 50 concepts reveals a significant divergence, explicitly exposing the previously discussed optimization dilemma among existing baselines. Methods prioritizing utility retention, such as GLoCE and HiRM, experience a severe collapse in unlearning efficacy, with their Unlearning Accuracy (UA) plummeting to 8.4\% and 23.2\%, respectively. Conversely, approaches enforcing aggressive erasure, including ESD, RECE, and ANT, virtually obliterate the foundational generative priors of the model, reducing Retention Accuracy (RA) to near zero. In contrast, KSCU strictly maintains exceptional erasure precision (99.52\% UA) even under challenging multi-concept interference with a 26.40\% RA. While naturally lower than its single-concept baseline, this represents the highest generative retention among all functionally successful unlearning methods, substantially surpassing competitors such as RECE (1.20\%) and ESD (0.36\%). These findings underscore the unparalleled stability and reliability of KSCU across scalable concept erasure tasks, with details provided in Appendix B.

\subsection{Ablation Study}

\begin{table}[!t]
\centering
\renewcommand{\arraystretch}{1.25}
\setlength{\tabcolsep}{3pt}
\caption{Ablation Study: Impact of the Key Step Table configuration (varying start step $S$, end step $E$, and interval $L$), Key Step Unlearning Optimization ($\lambda_1, \lambda_2$), and Prompt Augmentation (PA) on KSCU's performance.}
\begin{tabular}{@{}ccccc@{}}
\hline
\textbf{Method} & \textbf{UA(\%)$\uparrow$}  & \textbf{UDA(\%)$\downarrow$} & \textbf{FID-30k$\downarrow$} & \textbf{Time(s)$\downarrow$} \\
\hline
\small KSCU early 70\% steps  &91.1&59.8&15.6 &{541}\\
\small KSCU early 80\% steps &92.1&58.1&{14.1}&624\\
\small KSCU shrink $E$    & 88.5&89.3&15.3&570\\
\small KSCU last 60\% steps  &79.7 &85.3 &15.7&{498}\\
\small KSCU last 80\% steps   &{96.1}&{48.1} &14.3&628 \\
\small KSCU all steps    &96.0&54.5 &18.8&782\\
\hdashline
\small KSCU iter=500   &93.2& 50.8& 14.1 &367 \\
\small KSCU iter=1000  &96.7&47.2 &14.5&754\\
\hdashline
\small KSCU repeat   &96.0&54.6&15.7&2291\\
\small KSCU random   &92.9&51.3&14.9&2297\\
\hdashline
\small KSCU $\lambda_1$ = 0  &93.6&54.9&{13.5}&-\\
\small KSCU $\alpha$ = $2\times 10^{-4}$  & 96.9&44.2&18.4 &-\\
\small KSCU $\alpha$ = $2\times 10^{-3}$   &97.2&33.4&23.0&-\\
\small KSCU $\lambda_2$ = 0 &92.2&57.3&{14.1}&-\\
\small KSCU $\lambda_2 = 10^{-3}$  &97.3&30.1&16.3&-\\
\hdashline
\small KSCU w/o PA   &92.5&51.8 &14.5&-\\
\hdashline
\small KSCU (last 70\% steps) & {96.5} &{47.2} &{14.1}&546 \\
\hline
\end{tabular}
\label{tab:3}
\end{table} 

In this section, we systematically evaluate the three core components of KSCU: the Key Step Table, the Key Step Unlearning Optimization, and Prompt Augmentation.

To investigate the impact of the Key Step Table, we manipulate the temporal boundaries ($S$ and $E$) while controlling the step interval $L$ to ensure comparable computational budgets. As shown in Table \ref{tab:3}, shifting the unlearning focus toward the later stages of the denoising trajectory (e.g., comparing the early 70\% against the last 70\% of steps) consistently yields superior unlearning accuracy and adversarial robustness. This aligns with diffusion model mechanics, where early denoising steps primarily govern coarse spatial layouts, while later steps inject the fine-grained semantic details crucial for specific concept representations. 
\textcolor{black}{Crucially, the ending step $E$ must necessarily coincide with the absolute final step of the reverse process ($t \rightarrow 0$). As demonstrated by the ``KSCU shrink $E$'' setting, prematurely halting the intervention removes continuous gradient constraints during the final refinement phase, causing a drastic drop in unlearning accuracy (UA drops to 88.5\%). This occurs due to a ``concept re-synthesis'' phenomenon, where high-frequency features successfully suppressed in earlier intervals are actively re-synthesized by the model's latent generative priors when left unconstrained at the very end.}

\textcolor{black}{Furthermore, the specific starting thresholds evaluated here (e.g., last 60\%, last 70\%, last 80\%, and all steps) are not arbitrary hyperparameter searches, but serve as representative ``macro-sampling anchors'' to empirically validate our theoretical frequency destruction and emergence patterns. Specifically, the unlearning accuracies across the last 70\%, last 80\%, and all steps (100\%) are highly comparable (around 96\%), revealing that denoising steps prior to the 70\% mark contribute virtually nothing to erasing target high-frequency semantics. Instead, indiscriminately fine-tuning them introduces unnecessary global structural interference, evidenced by the FID degrading from 14.1 to 18.8. Conversely, when the interval is shrunk to the last 60\%, the UA drops drastically by 16.8\%. This prominent boundary transition effect decisively proves that the interval between the last 60\% and 70\% steps serves as the critical active region where the fine-grained high-frequency details of the target semantics intensely emerge in the latent space.} Consequently, targeted optimization across the last 70\% of steps establishes the optimal trade-off between computational efficiency, erasure precision, and retainability.

To verify whether the critical step table simply reduces the number of iterations to improve retention, we conducted ablation experiment on the total number of training iterations.
Prematurely halting the optimization (iter=500) fails to enforce thorough semantic erasure, dropping the UA to 93.2\% while leaving the FID stagnant at 14.1. Conversely, overextending the training (iter=1000) yields negligible unlearning gains but actively degrades generative retainability. This strictly confirms that operating within the mathematically identified active region, rather than simply restricting iteration counts, is the fundamental driver of performance. \textcolor{black}{Furthermore, to validate the necessity of our sequential step scheduling strategy, we benchmark our online learning formulation against unstructured alternatives: repeatedly sampling a fixed interval (repeat) and randomly selecting timesteps (random). Within an online learning paradigm, the optimization order is critical because the reverse diffusion process inherently relies on strict sequential trajectory dependency. Consequently, random sampling mathematically invalidates prior parameter updates by introducing unpredictable trajectory gaps, which causes a significant drop in unlearning efficacy (UA down to 92.9\%). Meanwhile, repeated sampling overfits to a narrow temporal window, triggering structural degradation. Our structured, forward-shifting step scheduling systematically circumvents these destructive interferences. By guaranteeing that parameter updates constructively accumulate along the continuous trajectory, it ensures consistent erasure precision alongside robust structural preservation.}

We next ablate the Key Step Unlearning Optimization by systematically examining the loss weighting parameters, $\lambda_1$ and $\lambda_2$. A core innovation of KSCU is the timestep-dependent dynamic scheduling of the unconditional predicted noise optimization, formulated as $\lambda_1 = (1000 - t) \times \alpha$. This design assigns smaller penalty weights during early denoising steps (where $t$ is large) to protect the global structural layout from disruption, while applying heavier penalties during later steps to thoroughly eradicate fine-grained semantic details. As shown in Table \ref{tab:3}, completely disabling this term ($\lambda_1=0$) yields the best generative retainability (FID 13.5) but noticeably cripples unlearning efficacy (UA 93.6\%) and adversarial robustness (UDA 54.9\%). Conversely, increasing the base scaling factor $\alpha$ to $2\times 10^{-4}$ and further to $2\times 10^{-3}$ pushes the UA up to 97.2\%, but progressively corrupts the generative quality, driving the FID up to 18.4 and 23.0, respectively. This confirms that the dynamic schedule requires a carefully calibrated scaling factor to prevent catastrophic over-unlearning.

Similar inherent trade-offs are observed when evaluating the regularization term $\lambda_2$. Removing it entirely ($\lambda_2=0$) severely degrades the overall unlearning performance, dropping UA to 92.2\% and UDA to 57.3\%. Conversely, applying a strong static weight ($\lambda_2=10^{-3}$) aggressively boosts erasure precision (UA 97.3\%) and adversarial robustness (UDA 30.1\%), but introduces collateral damage to general image quality, raising the FID to 16.3. Ultimately, the standard KSCU configuration establishes an optimal equilibrium between these dynamic and static objectives, achieving a superior UA of 96.5\% while firmly locking the FID at 14.1.

Finally, removing the Prompt Augmentation module leads to a discernible drop in UA (92.5\%) and a significant surge in adversarial vulnerability, with UDA rising to 51.8\%. Notably, the FID remains strictly unaffected at 14.5. This confirms that prompt augmentation is a highly lightweight yet strictly necessary mechanism. It substantially fortifies the model's resilience against complex or adversarial prompt variations without imposing any collateral damage on generative quality.
\subsection{Versatility of the Key Step Plugin}

\begin{table}[!t]
\centering
\caption{Evaluation of the Key Step Plugin on the Nudity task. Integrating the proposed Key Step optimization into existing baselines (denoted with the $key$ subscript) universally improves training speed, generative retainability (FID), and robustness (UDA).}
\begin{tabular}{lcccc}
\hline
\textbf{\multirow{2}{*}{Method}} & \multicolumn{2}{c}{\textbf{Effective}} & \multicolumn{1}{c}{\textbf{Retain}} & \multicolumn{1}{c}{\textbf{Speed}} \\
\cline{2-5}
  &\multicolumn{1}{c}{\textbf{UA(\%)$\uparrow$}}  &\multicolumn{1}{c}{\textbf{UDA(\%)$\downarrow$}} & \multicolumn{1}{c}{\textbf{FID-30k$\downarrow$}} &\multicolumn{1}{c}{\textbf{Time(s)$\downarrow$}} \\
\hline
ESD      & 88.2 & 73.7 & 15.4 & 2557 \\
$\text{ESD}_\text{key}$  & 90.5 & 62.7 & 15.0 & 2289 \\
\hdashline
EDiff    & 93.9 & 52.5 & 19.8 & 1914 \\
$\text{EDiff}_\text{key}$& 93.1 & 49.3 & 17.5 & 1399 \\
\hdashline
SHS      & 84.3 & 95.5 & 18.1 & 817  \\
$\text{SHS}_\text{key}$  & 83.1 & 95.2 & 15.2 & 627  \\
\hline
\end{tabular}
\label{tab:transfer}
\end{table}

To further validate the universal applicability and theoretical soundness of our strategy, we evaluate the Key Step mechanism as an independent plugin. Specifically, we extract the core Key Step Table algorithm, which strictly restricts unlearning fine-tuning to the later and semantic-heavy denoising steps, and integrate it into three existing state-of-the-art baselines: ESD, EDiff, and SHS. These enhanced variants are denoted as $\text{ESD}_\text{key}$, $\text{EDiff}_\text{key}$, and $\text{SHS}_\text{key}$.

As demonstrated in Table \ref{tab:transfer}, functioning as a plug-and-play plugin yields universal and substantial improvements across all three diverse baselines. Most notably, restricting the optimization steps drastically reduces computational overhead. Furthermore, by shielding the early layout-formation steps from indiscriminate optimization, the plugin variants significantly mitigate collateral damage to the fundamental generative priors of the models. This preservation of structural integrity is evidenced by consistent reductions in FID across the board (e.g., $\text{EDiff}_\text{key}$ improves FID from 19.8 to 17.5, and $\text{SHS}_\text{key}$ drops from 18.1 to 15.2).

Crucially, these simultaneous gains in training speed and image quality do not compromise unlearning efficacy. The modified baselines either strictly maintain or explicitly enhance their target erasure precision and adversarial robustness. For instance, $\text{ESD}_\text{key}$ not only generates higher-quality images faster than the original ESD but also improves the Unlearn Accuracy from 88.2\% to 90.5\% while lowering the adversarial vulnerability (UDA) from 73.7\% to 62.7\%.

\section{Conclusion}
In this work, we introduce Key Step Concept Unlearning (KSCU), a theoretically grounded framework that strictly isolates parameter optimization to an intermediate active region of the denoising trajectory. By aligning fine-tuning with periods of high gradient sensitivity and low trajectory deviation, KSCU successfully prevents the structural collapse typically caused by early-step interference. Extensive empirical evaluations across nudity, style, class, and mass concept erasure demonstrate that KSCU consistently achieves the optimal unlearning-retainment trade-off.
Future work will explore optimizing dynamic concept decoupling strategies during the iterative process to further mitigate KSCU's retainability degradation in extreme mass unlearning scenarios involving hundreds of concurrent concepts.



\bibliographystyle{IEEEtran}
\bibliography{aaai2026}

@inproceedings{GLoCE,
  title={Localized concept erasure for text-to-image diffusion models using training-free gated low-rank adaptation},
  author={Lee, Byung Hyun and Lim, Sungjin and Chun, Se Young},
  booktitle={Proceedings of the Computer Vision and Pattern Recognition Conference},
  pages={18596--18606},
  year={2025}
}

@article{HiRM,
  title={Localized Concept Erasure in Text-to-Image Diffusion Models via High-Level Representation Misdirection},
  author={Lee, Uichan and Kim, Jeonghyeon and Hwang, Sangheum},
  journal={arXiv preprint arXiv:2602.19631},
  year={2026}
}

@article{li2026aegis,
  title={AEGIS: Adversarial Target-Guided Retention-Data-Free Robust Concept Erasure from Diffusion Models},
  author={Li, Fengpeng and Li, Kemou and Wang, Qizhou and Han, Bo and Zhou, Jiantao},
  journal={arXiv preprint arXiv:2602.06771},
  year={2026}
}

@article{hertz2022prompt,
  title={Prompt-to-prompt image editing with cross attention control},
  author={Hertz, Amir and Mokady, Ron and Tenenbaum, Jay and Aberman, Kfir and Pritch, Yael and Cohen-Or, Daniel},
  journal={arXiv preprint arXiv:2208.01626},
  year={2022}
}

@inproceedings{shilocalized,
  title={Localized-Attention-Guided Concept Erasure for Text-to-Image Diffusion Models},
  author={Shi, Zhuan and Farashah, Alireza Dehghanpour and de Vries, Rik and Farnadi, Golnoosh},
  booktitle={Generative and Protective AI for Content Creation}
}

@inproceedings{si2024freeu,
  title={Freeu: Free lunch in diffusion u-net},
  author={Si, Chenyang and Huang, Ziqi and Jiang, Yuming and Liu, Ziwei},
  booktitle={Proceedings of the IEEE/CVF conference on computer vision and pattern recognition},
  pages={4733--4743},
  year={2024}
}

@inproceedings{Conceptreplacer,
  title={Concept replacer: Replacing sensitive concepts in diffusion models via precision localization},
  author={Zhang, Lingyun and Xie, Yu and Fu, Yanwei and Chen, Ping},
  booktitle={Proceedings of the Computer Vision and Pattern Recognition Conference},
  pages={8172--8181},
  year={2025}
}

@inproceedings{Conceptsliders,
  title={Concept sliders: Lora adaptors for precise control in diffusion models},
  author={Gandikota, Rohit and Materzy{\'n}ska, Joanna and Zhou, Tingrui and Torralba, Antonio and Bau, David},
  booktitle={European Conference on Computer Vision},
  pages={172--188},
  year={2024},
  organization={Springer}
}

@article{sun2026lure,
  title={LURE: Latent Space Unblocking for Multi-Concept Reawakening in Diffusion Models},
  author={Sun, Mengyu and Yang, Ziyuan and Teoh, Andrew Beng Jin and Liu, Junxu and Hu, Haibo and Zhang, Yi},
  journal={arXiv preprint arXiv:2601.14330},
  year={2026}
}

@article{tu2026mass,
  title={Mass Concept Erasure in Diffusion Models with Concept Hierarchy},
  author={Tu, Jiahang and Li, Ye and Wu, Yiming and Zhao, Hanbin and Zhang, Chao and Qian, Hui},
  journal={arXiv preprint arXiv:2601.03305},
  year={2026}
}

@inproceedings{ren2025six,
  title={Six-cd: Benchmarking concept removals for text-to-image diffusion models},
  author={Ren, Jie and Chen, Kangrui and Cui, Yingqian and Zeng, Shenglai and Liu, Hui and Xing, Yue and Tang, Jiliang and Lyu, Lingjuan},
  booktitle={Proceedings of the Computer Vision and Pattern Recognition Conference},
  pages={28769--28778},
  year={2025}
}

@article{zhang2026differential,
  title={Differential Vector Erasure: Unified Training-Free Concept Erasure for Flow Matching Models},
  author={Zhang, Zhiqi and Zhong, Xinhao and Sun, Yi and Sun, Shuoyang and Chen, Bin and Xia, Shu-Tao and Wang, Xuan},
  journal={arXiv preprint arXiv:2602.01089},
  year={2026}
}

@inproceedings{AdaVD,
  title={Precise, fast, and low-cost concept erasure in value space: Orthogonal complement matters},
  author={Wang, Yuan and Li, Ouxiang and Mu, Tingting and Hao, Yanbin and Liu, Kuien and Wang, Xiang and He, Xiangnan},
  booktitle={2025 IEEE/CVF Conference on Computer Vision and Pattern Recognition (CVPR)},
  pages={28759--28768},
  year={2025},
  organization={IEEE}
}

@article{esd,
  title={Erasing concepts from diffusion models},
  author={Gandikota, Rohit and Materzynska, Joanna and Fiotto-Kaufman, Jaden and Bau, David},
  journal={Proceedings of the IEEE/CVF International Conference on Computer Vision},
  pages={2426--2436},
  year={2023}
}

@article{unlearndiffatk,
  title={To generate or not? safety-driven unlearned diffusion models are still easy to generate unsafe images... for now},
  author={Zhang, Yimeng and Jia, Jinghan and Chen, Xin and Chen, Aochuan and Zhang, Yihua and Liu, Jiancheng and Ding, Ke and Liu, Sijia},
  journal={European Conference on Computer Vision},
  pages={385--403},
  year={2024},
  organization={Springer}
}

@article{fmn,
  title={Forget-me-not: Learning to forget in text-to-image diffusion models},
  author={Zhang, Gong and Wang, Kai and Xu, Xingqian and Wang, Zhangyang and Shi, Humphrey},
  journal={Proceedings of the IEEE/CVF conference on computer vision and pattern recognition},
  pages={1755--1764},
  year={2024}
}

@article{ca,
  title={Ablating concepts in text-to-image diffusion models},
  author={Kumari, Nupur and Zhang, Bingliang and Wang, Sheng-Yu and Shechtman, Eli and Zhang, Richard and Zhu, Jun-Yan},
  journal={Proceedings of the IEEE/CVF International Conference on Computer Vision},
  pages={22691--22702},
  year={2023}
}

@article{ringabell,
  title={Ring-a-bell! how reliable are concept removal methods for diffusion models?},
  author={Tsai, Yu-Lin and Hsu, Chia-Yi and Xie, Chulin and Lin, Chih-Hsun and Chen, Jia-You and Li, Bo and Chen, Pin-Yu and Yu, Chia-Mu and Huang, Chun-Ying},
  journal={arXiv preprint arXiv:2310.10012},
  year={2023}
}

@inproceedings{SDErasure,
  title={SDErasure: Concept-Specific Trajectory Shifting for Concept Erasure via Adaptive Diffusion Classifier},
  author={Miao, Fengyuan and Fang, Shancheng and Yu, Lingyun and Qu, Yadong and Sun, Yuhao and Wang, Xiaorui and Xie, Hongtao},
  booktitle={The Fourteenth International Conference on Learning Representations}
}

@article{shs,
  title={Scissorhands: Scrub data influence via connection sensitivity in networks},
  author={Wu, Jing and Harandi, Mehrtash},
  journal={European Conference on Computer Vision},
  pages={367--384},
  year={2024},
  organization={Springer}
}

@article{wu2024erasediff,
  title={Erasing undesirable influence in diffusion models},
  author={Wu, Jing and Le, Trung and Hayat, Munawar and Harandi, Mehrtash},
  booktitle={Proceedings of the Computer Vision and Pattern Recognition Conference},
  year={2025}
}

@article{ant,
  title={Set you straight: Auto-steering denoising trajectories to sidestep unwanted concepts},
  author={Li, Leyang and Lu, Shilin and Ren, Yan and Kong, Adams Wai-Kin},
  journal={arXiv preprint arXiv:2504.12782},
  year={2025}
}

@article{speed,
  title={Speed: Scalable, precise, and efficient concept erasure for diffusion models},
  author={Li, Ouxiang and Wang, Yuan and Hu, Xinting and Jiang, Houcheng and Liang, Tao and Hao, Yanbin and Ma, Guojun and Feng, Fuli},
  journal={arXiv preprint arXiv:2503.07392},
  year={2025}
}

@inproceedings{mace,
  title={
  els},
  author={Lu, Shilin and Wang, Zilan and Li, Leyang and Liu, Yanzhu and Kong, Adams Wai-Kin},
  booktitle={Proceedings of the IEEE/CVF Conference on Computer Vision and Pattern Recognition},
  pages={6430--6440},
  year={2024}
}

@article{goodfellow2020gan,
  title={Generative adversarial networks},
  author={Goodfellow, Ian and Pouget-Abadie, Jean and Mirza, Mehdi and Xu, Bing and Warde-Farley, David and Ozair, Sherjil and Courville, Aaron and Bengio, Yoshua},
  journal={Communications of the ACM},
  volume={63},
  number={11},
  pages={139--144},
  year={2020},
  publisher={ACM New York, NY, USA}
}

@article{poole2022dreamfusion,
  title={Dreamfusion: Text-to-3d using 2d diffusion},
  author={Poole, Ben and Jain, Ajay and Barron, Jonathan T and Mildenhall, Ben},
  journal={arXiv preprint arXiv:2209.14988},
  year={2022}
}

@inproceedings{blattmann2023align,
  title={Align your latents: High-resolution video synthesis with latent diffusion models},
  author={Blattmann, Andreas and Rombach, Robin and Ling, Huan and Dockhorn, Tim and Kim, Seung Wook and Fidler, Sanja and Kreis, Karsten},
  booktitle={Proceedings of the IEEE/CVF conference on computer vision and pattern recognition},
  pages={22563--22575},
  year={2023}
}

@article{kingma2013vae,
  title={Auto-encoding variational bayes},
  author={Kingma, Diederik P and Welling, Max},
  journal={arXiv preprint arXiv:1312.6114},
  year={2013}
}

@article{rece,
  title={Reliable and efficient concept erasure of text-to-image diffusion models},
  author={Gong, Chao and Chen, Kai and Wei, Zhipeng and Chen, Jingjing and Jiang, Yu-Gang},
  journal={European Conference on Computer Vision},
  pages={73--88},
  year={2024},
  organization={Springer}
}

@article{fan2023salun,
  title={Salun: Empowering machine unlearning via gradient-based weight saliency in both image classification and generation},
  author={Fan, Chongyu and Liu, Jiancheng and Zhang, Yihua and Wong, Eric and Wei, Dennis and Liu, Sijia},
  journal={arXiv preprint arXiv:2310.12508},
  year={2023}
}

@article{uce,
  title={Unified concept editing in diffusion models},
  author={Gandikota, Rohit and Orgad, Hadas and Belinkov, Yonatan and Materzy{\'n}ska, Joanna and Bau, David},
  journal={Proceedings of the IEEE/CVF Winter Conference on Applications of Computer Vision},
  pages={5111--5120},
  year={2024}
}

@article{advunlearn,
  title={Defensive unlearning with adversarial training for robust concept erasure in diffusion models},
  author={Zhang, Yimeng and Chen, Xin and Jia, Jinghan and Zhang, Yihua and Fan, Chongyu and Liu, Jiancheng and Hong, Mingyi and Ding, Ke and Liu, Sijia},
  journal={Advances in Neural Information Processing Systems},
  volume={37},
  pages={36748--36776},
  year={2025}
}

@article{p4d,
  title={Prompting4debugging: Red-teaming text-to-image diffusion models by finding problematic prompts},
  author={Chin, Zhi-Yi and Jiang, Chieh-Ming and Huang, Ching-Chun and Chen, Pin-Yu and Chiu, Wei-Chen},
  journal={arXiv preprint arXiv:2309.06135},
  year={2023}
}

@article{zhang2024unlearncanvas,
  title={UnlearnCanvas: Stylized Image Dataset for Enhanced Machine Unlearning Evaluation in Diffusion Models},
  author={Zhang, Yihua and Fan, Chongyu and Zhang, Yimeng and Yao, Yuguang and Jia, Jinghan and Liu, Jiancheng and Zhang, Gaoyuan and Liu, Gaowen and Kompella, Ramana Rao and Liu, Xiaoming and others},
  journal={arXiv preprint arXiv:2402.11846},
  year={2024}
}

@article{i2p,
  title={Safe latent diffusion: Mitigating inappropriate degeneration in diffusion models},
  author={Schramowski, Patrick and Brack, Manuel and Deiseroth, Bj{\"o}rn and Kersting, Kristian},
  journal={Proceedings of the IEEE/CVF Conference on Computer Vision and Pattern Recognition},
  pages={22522--22531},
  year={2023}
}

@article{ldm,
  title={High-resolution image synthesis with latent diffusion models},
  author={Rombach, Robin and Blattmann, Andreas and Lorenz, Dominik and Esser, Patrick and Ommer, Bj{\"o}rn},
  journal={Proceedings of the IEEE/CVF conference on computer vision and pattern recognition},
  pages={10684--10695},
  year={2022}
}

@article{kingma2021variational,
  title={Variational diffusion models},
  author={Kingma, Diederik and Salimans, Tim and Poole, Ben and Ho, Jonathan},
  journal={Advances in neural information processing systems},
  volume={34},
  pages={21696--21707},
  year={2021}
}

@article{kawar2023imagic,
  title={Imagic: Text-based real image editing with diffusion models},
  author={Kawar, Bahjat and Zada, Shiran and Lang, Oran and Tov, Omer and Chang, Huiwen and Dekel, Tali and Mosseri, Inbar and Irani, Michal},
  journal={Proceedings of the IEEE/CVF conference on computer vision and pattern recognition},
  pages={6007--6017},
  year={2023}
}

@article{dbg,
  title={Diffusion models beat gans on image synthesis},
  author={Dhariwal, Prafulla and Nichol, Alexander},
  journal={Advances in neural information processing systems},
  volume={34},
  pages={8780--8794},
  year={2021}
}

@article{ho2022classifierfree,
  title={Classifier-free diffusion guidance},
  author={Ho, Jonathan and Salimans, Tim},
  journal={arXiv preprint arXiv:2207.12598},
  year={2022}
}

@article{rando2022redfilter,
  title={Red-teaming the stable diffusion safety filter},
  author={Rando, Javier and Paleka, Daniel and Lindner, David and Heim, Lennart and Tram{\`e}r, Florian},
  journal={arXiv preprint arXiv:2210.04610},
  year={2022}
}

@article{laion,
  title={Laion-400m: Open dataset of clip-filtered 400 million image-text pairs},
  author={Schuhmann, Christoph and Vencu, Richard and Beaumont, Romain and Kaczmarczyk, Robert and Mullis, Clayton and Katta, Aarush and Coombes, Theo and Jitsev, Jenia and Komatsuzaki, Aran},
  journal={arXiv preprint arXiv:2111.02114},
  year={2021}
}

@article{carlini2023extracting,
  title={Extracting training data from diffusion models},
  author={Carlini, Nicolas and Hayes, Jamie and Nasr, Milad and Jagielski, Matthew and Sehwag, Vikash and Tramer, Florian and Balle, Borja and Ippolito, Daphne and Wallace, Eric},
  journal={32nd USENIX Security Symposium (USENIX Security 23)},
  pages={5253--5270},
  year={2023}
}

@article{ramesh2022hierarchical,
  title={Hierarchical text-conditional image generation with clip latents},
  author={Ramesh, Aditya and Dhariwal, Prafulla and Nichol, Alex and Chu, Casey and Chen, Mark},
  journal={arXiv preprint arXiv:2204.06125},
  volume={1},
  number={2},
  pages={3},
  year={2022}
}

@article{yang2023diffusion,
  title={Diffusion probabilistic model made slim},
  author={Yang, Xingyi and Zhou, Daquan and Feng, Jiashi and Wang, Xinchao},
    journal={Proceedings of the IEEE/CVF Conference on computer vision and pattern recognition},
  pages={22552--22562},
  year={2023}
}

@article{cao2025diffstereo,
  title={DiffStereo: High-Frequency Aware Diffusion Model for Stereo Image Restoration},
  author={Cao, Huiyun and Shi, Yuan and Xia, Bin and Jin, Xiaoyu and Yang, Wenming},
  journal={arXiv preprint arXiv:2501.10325},
  year={2025}
}

@article{chen2025safe,
  title={Safe and Reliable Diffusion Models via Subspace Projection},
  author={Chen, Huiqiang and Zhu, Tianqing and Wang, Linlin and Yu, Xin and Gao, Longxiang and Zhou, Wanlei},
  journal={arXiv preprint arXiv:2503.16835},
  year={2025}
}

@article{xue2025crce,
  title={CRCE: Coreference-Retention Concept Erasure in Text-to-Image Diffusion Models},
  author={Xue, Yuyang and Moroshko, Edward and Chen, Feng and McDonagh, Steven and Tsaftaris, Sotirios A},
  journal={arXiv preprint arXiv:2503.14232},
  year={2025}
}

@article{ho2020denoising,
  title={Denoising diffusion probabilistic models},
  author={Ho, Jonathan and Jain, Ajay and Abbeel, Pieter},
  journal={Advances in neural information processing systems},
  volume={33},
  pages={6840--6851},
  year={2020}
}

@article{song2020score,
  title={Score-based generative modeling through stochastic differential equations},
  author={Song, Yang and Sohl-Dickstein, Jascha and Kingma, Diederik P and Kumar, Abhishek and Ermon, Stefano and Poole, Ben},
  journal={arXiv preprint arXiv:2011.13456},
  year={2020}
}

@article{fid,
  title={Gans trained by a two time-scale update rule converge to a local nash equilibrium},
  author={Heusel, Martin and Ramsauer, Hubert and Unterthiner, Thomas and Nessler, Bernhard and Hochreiter, Sepp},
  journal={Advances in neural information processing systems},
  volume={30},
  year={2017}
}

@article{coco,
  title={Microsoft coco: Common objects in context},
  author={Lin, Tsung-Yi and Maire, Michael and Belongie, Serge and Hays, James and Perona, Pietro and Ramanan, Deva and Doll{\'a}r, Piotr and Zitnick, C Lawrence},
  journal={Computer vision--ECCV 2014: 13th European conference, zurich, Switzerland, September 6-12, 2014, proceedings, part v 13},
  pages={740--755},
  year={2014},
  organization={Springer}
}

@article{ncsm,
  title={Generative modeling by estimating gradients of the data distribution},
  author={Song, Yang and Ermon, Stefano},
  journal={Advances in neural information processing systems},
  volume={32},
  year={2019}
}

@article{lu2022dpm,
  title={Dpm-solver: A fast ode solver for diffusion probabilistic model sampling in around 10 steps},
  author={Lu, Cheng and Zhou, Yuhao and Bao, Fan and Chen, Jianfei and Li, Chongxuan and Zhu, Jun},
  journal={Advances in Neural Information Processing Systems},
  volume={35},
  pages={5775--5787},
  year={2022}
}

@online{chatgpt,
  author    = {OpenAI},
  organization = {OpenAI},
  title     = {ChatGPT: A Conversational AI Model},
  year      = {2025},
  url       = {https://openai.com/chatgpt},
  note      = {Accessed: 2025-03}
}

@online{deepseek,
  author       = {DeepSeek},
  organization = {DeepSeek AI},
  title        = {DeepSeek Chat: A Conversational LLM},
  year         = {2025},
  url          = {https://chat.deepseek.com/},
  note         = {Accessed: 2025-03}
}

@online{stability,
  author       = {Stability AI},
  organization = {Stability AI Ltd.},
  title        = {Stability AI: An image generation tool},
  year         = {2025},
  url          = {https://stability.ai/},
  note         = {Accessed: 2025-03}
}

@online{nudenet,
  author       = {Bedapudi Praneeth},
  title        = {NudeNet: lightweight Nudity detection},
  year         = {2025},
  howpublished = {GitHub repository},
  url          = {https://github.com/notAI-tech/NudeNet},
  note         = {Accessed: 2025-03}
}

@article{ddim,
  title={Denoising diffusion implicit models},
  author={Song, Jiaming and Meng, Chenlin and Ermon, Stefano},
  journal={arXiv preprint arXiv:2010.02502},
  year={2020}
}

@article{liu2024sora,
  title={Sora: A review on background, technology, limitations, and opportunities of large vision models},
  author={Liu, Yixin and Zhang, Kai and Li, Yuan and Yan, Zhiling and Gao, Chujie and Chen, Ruoxi and Yuan, Zhengqing and Huang, Yue and Sun, Hanchi and Gao, Jianfeng and others},
  journal={arXiv preprint arXiv:2402.17177},
  year={2024}
}

@article{meng2022locating,
  title={Locating and editing factual associations in gpt},
  author={Meng, Kevin and Bau, David and Andonian, Alex and Belinkov, Yonatan},
  journal={Advances in neural information processing systems},
  volume={35},
  pages={17359--17372},
  year={2022}
}

@article{liu2022compositional,
  title={Compositional visual generation with composable diffusion models},
  author={Liu, Nan and Li, Shuang and Du, Yilun and Torralba, Antonio and Tenenbaum, Joshua B},
  journal={European Conference on Computer Vision},
  pages={423--439},
  year={2022},
  organization={Springer}
}

@article{saharia2022photorealistic,
  title={Photorealistic text-to-image diffusion models with deep language understanding},
  author={Saharia, Chitwan and Chan, William and Saxena, Saurabh and Li, Lala and Whang, Jay and Denton, Emily L and Ghasemipour, Kamyar and Gontijo Lopes, Raphael and Karagol Ayan, Burcu and Salimans, Tim and others},
  journal={Advances in neural information processing systems},
  volume={35},
  pages={36479--36494},
  year={2022}
}

@article{shan2023glaze,
  title={Glaze: Protecting artists from style mimicry by $\{$Text-to-Image$\}$ models},
  author={Shan, Shawn and Cryan, Jenna and Wenger, Emily and Zheng, Haitao and Hanocka, Rana and Zhao, Ben Y},
  journal={32nd USENIX Security Symposium (USENIX Security 23)},
  pages={2187--2204},
  year={2023}
}

@article{changpinyo2021conceptual,
  title={Conceptual 12m: Pushing web-scale image-text pre-training to recognize long-tail visual concepts},
  author={Changpinyo, Soravit and Sharma, Piyush and Ding, Nan and Soricut, Radu},
  journal={Proceedings of the IEEE/CVF conference on computer vision and pattern recognition},
  pages={3558--3568},
  year={2021}
}

@article{ramesh2021zero,
  title={Zero-shot text-to-image generation},
  author={Ramesh, Aditya and Pavlov, Mikhail and Goh, Gabriel and Gray, Scott and Voss, Chelsea and Radford, Alec and Chen, Mark and Sutskever, Ilya},
  journal={International conference on machine learning},
  pages={8821--8831},
  year={2021},
  organization={Pmlr}
}

@article{park2024direct,
  title={Direct unlearning optimization for robust and safe text-to-image models},
  author={Park, Yong-Hyun and Yun, Sangdoo and Kim, Jin-Hwa and Kim, Junho and Jang, Geonhui and Jeong, Yonghyun and Jo, Junghyo and Lee, Gayoung},
  journal={arXiv preprint arXiv:2407.21035},
  year={2024}
}

@article{sclocchi2025phase,
  title={A phase transition in diffusion models reveals the hierarchical nature of data},
  author={Sclocchi, Antonio and Favero, Alessandro and Wyart, Matthieu},
  journal={Proceedings of the National Academy of Sciences},
  volume={122},
  number={1},
  pages={e2408799121},
  year={2025},
  publisher={National Academy of Sciences}
}

@inproceedings{lee2025beta,
  title={Beta Sampling is All You Need: Efficient Image Generation Strategy for Diffusion Models Using Stepwise Spectral Analysis},
  author={Lee, Haeil and Lee, Hansang and Gye, Seoyeon and Kim, Junmo},
  booktitle={2025 IEEE/CVF Winter Conference on Applications of Computer Vision (WACV)},
  pages={4215--4224},
  year={2025},
  organization={IEEE}
}
\appendices
\setcounter{equation}{0}
\renewcommand{\theequation}{A\arabic{equation}}
\section{Information-Theoretic Proof of Sequential Frequency Destruction}
\label{subsec:sequential_destruction_info}

To rigorously formalize the concept of "information destruction" and prove that high-frequency semantic details are lost strictly earlier than low-frequency global structures, we analyze the forward diffusion process from an information-theoretic perspective.

\textbf{Assumption 1 (Second-Order Stationary Random Field).}
We model the initial clean image $\mathbf{x}_0$ as a second-order stationary Gaussian random field defined on the continuous spatial domain $\mathbb{R}^d$. Under this assumption, the spatial Fourier transform diagonalizes the covariance operator, meaning the frequency components $\hat{\mathbf{x}}_0(\omega)$ are statistically uncorrelated. This allows us to analyze the information degradation independently frequency-by-frequency.

\textbf{Assumption 2 (Monotone Radially Averaged Spectral Decay).}
While the raw 2D power spectrum of a natural image contains local fluctuations, its expected \textit{radially averaged} Power Spectral Density (PSD), denoted as $\mathcal{S}_{\mathbf{x}_0}(\|\omega\|)$, follows a power-law decay. We assume $\mathcal{S}_{\mathbf{x}_0}(\|\omega\|)$ is a strictly monotonically decreasing function with respect to the frequency magnitude $\|\omega\|$.

\textbf{Lemma 1 (Global Ordering of SNR).}
For any two spatial frequencies $\omega_1$ and $\omega_2$, if $\|\omega_1\| > \|\omega_2\|$, Assumption 2 strictly implies $\mathcal{S}_{\mathbf{x}_0}(\|\omega_1\|) < \mathcal{S}_{\mathbf{x}_0}(\|\omega_2\|)$. Since the global noise scaling factor $f(t) = \frac{\alpha(t)}{1 - \alpha(t)} > 0$, the frequency-dependent $\text{SNR}(\omega, t) = f(t)\mathcal{S}_{\mathbf{x}_0}(\|\omega\|)$ preserves this strict ordering:
\begin{equation}
\text{SNR}(\omega_1, t) < \text{SNR}(\omega_2, t), \quad \forall t \in (0, T).
\end{equation}

\textbf{Definition 1 (Information Destruction Time via Mutual Information).}
To define "information loss" strictly within the framework of Shannon Information Theory, we measure the Mutual Information (MI) between the initial frequency component $\hat{\mathbf{x}}_0(\omega)$ and its diffused counterpart $\hat{\mathbf{x}}_t(\omega)$. Since the forward diffusion process in the frequency domain acts as an Additive White Gaussian Noise (AWGN) channel, the mutual information is exactly formulated as:
\begin{equation}
\mathcal{I}(\hat{\mathbf{x}}_0(\omega); \hat{\mathbf{x}}_t(\omega)) = \frac{1}{2} \log \big( 1 + \text{SNR}(\omega, t) \big).
\end{equation}
Notice that as $\text{SNR} \to 0$ (equivalent to $\text{MMSE} \to 1$), the mutual information $\mathcal{I} \to 0$. We define the \textit{information destruction time} $t_\omega$ as the earliest time the mutual information drops below a negligible threshold $\epsilon > 0$, representing the state of statistical unrecoverability:
\begin{equation}
t_\omega := \inf \left\{ t \in (0, T] : \mathcal{I}(\hat{\mathbf{x}}_0(\omega); \hat{\mathbf{x}}_t(\omega)) \le \epsilon \right\}.
\end{equation}

At the critical time $t = t_\omega$, the boundary condition implies:
\begin{equation}
\frac{1}{2} \log \big( 1 + \text{SNR}(\omega, t_\omega) \big) = \epsilon \implies \text{SNR}(\omega, t_\omega) = e^{2\epsilon} - 1.
\end{equation}

Substituting $\text{SNR}(\omega, t) = f(t)\mathcal{S}_{\mathbf{x}_0}(\|\omega\|)$, we obtain the explicit equation for the global scaling factor at $t_\omega$:
\begin{equation}
f(t_\omega) = \frac{e^{2\epsilon} - 1}{\mathcal{S}_{\mathbf{x}_0}(\|\omega\|)}.
\end{equation}

\textbf{Theorem 1 (Sequential Destruction of Frequencies).}
For any two spatial frequencies $\omega_1$ and $\omega_2$, if $\|\omega_1\| > \|\omega_2\|$, the mutual information of the higher frequency component reaches the destruction threshold strictly earlier, i.e., $t_{\omega_1} < t_{\omega_2}$.

\textbf{Proof.}
By the definition of the standard variance-preserving noise schedule, the signal attenuation factor $\alpha(t)$ is continuous and strictly monotonically decreasing from $\alpha(0)=1$ to $\alpha(T) \approx 0$. Consequently, the function $f(t) = \frac{\alpha(t)}{1 - \alpha(t)}$ is a continuous and strictly monotonically decreasing bijection from $(0, T)$ to $(0, \infty)$.
Thus, its inverse function $f^{-1}: (0, \infty) \to (0, T)$ exists, is well-defined, and is strictly monotonically decreasing.

Assume $\epsilon$ is sufficiently small such that $\frac{e^{2\epsilon} - 1}{\mathcal{S}_{\mathbf{x}_0}(\|\omega\|)} \in (0, \infty)$ for all valid frequencies, ensuring the crossing time $t_\omega$ is strictly bounded within $(0, T)$. We can explicitly formulate the destruction time as:
\begin{equation}
t_\omega = f^{-1} \left( \frac{e^{2\epsilon} - 1}{\mathcal{S}_{\mathbf{x}_0}(\|\omega\|)} \right).
\end{equation}

From Lemma 1, $\|\omega_1\| > \|\omega_2\| \implies \mathcal{S}_{\mathbf{x}_0}(\|\omega_1\|) < \mathcal{S}_{\mathbf{x}_0}(\|\omega_2\|)$. Since both PSD values are positive, taking their reciprocals yields:
\begin{equation}
\frac{e^{2\epsilon} - 1}{\mathcal{S}_{\mathbf{x}_0}(\|\omega_1\|)} > \frac{e^{2\epsilon} - 1}{\mathcal{S}_{\mathbf{x}_0}(\|\omega_2\|)}.
\end{equation}

Because the inverse function $f^{-1}$ is strictly monotonically decreasing, applying $f^{-1}$ to both sides of the inequality strictly reverses the relation:
\begin{equation}
f^{-1} \left( \frac{e^{2\epsilon} - 1}{\mathcal{S}_{\mathbf{x}_0}(\|\omega_1\|)} \right) < f^{-1} \left( \frac{e^{2\epsilon} - 1}{\mathcal{S}_{\mathbf{x}_0}(\|\omega_2\|)} \right).
\end{equation}

Substituting the explicit formulation of the destruction time, we obtain exactly:
\begin{equation}
t_{\omega_1} < t_{\omega_2}.
\end{equation}
This rigorously establishes that high-frequency semantic details suffer complete information-theoretic destruction strictly earlier than low-frequency global structures during the forward diffusion process.

\section{Reverse Denoising Dynamics: The Sequential Emergence of Frequencies}
\label{subsec:reverse_emergence}

Having established the chronological order of information destruction in the forward process, we now analyze the reverse generative process. The goal is to rigorously prove that during generation (where time progresses backwards from $t=T$ to $t=0$), low-frequency global structures emerge strictly earlier than high-frequency semantic details.

\textbf{Definition 2 (Frequency Reconstruction Weight via Tweedie's Formula).}
At any reverse diffusion step $t$, the neural network denoiser estimates the clean data $\mathbf{x}_0$. According to Tweedie's formula, the Minimum Mean Square Error (MMSE) estimator in the frequency domain intrinsically acts as a Wiener filter. The effective reconstruction weight $H(\omega, t)$ applied to the frequency component $\omega$ is heavily dependent on the local SNR:
\begin{equation}
H(\omega, t) = \frac{\text{SNR}(\omega, t)}{1 + \text{SNR}(\omega, t)}.
\end{equation}
The weight $H(\omega, t) \in [0, 1)$ dictates the network's confidence in restoring that specific frequency. When $H \approx 0$, the frequency is entirely suppressed (treated as noise); as $H$ grows, the frequency is actively reconstructed.

\textbf{Definition 3 (Information Emergence Time).}
In the reverse process, time evolves backwards from $T$ down to $0$. We define a reconstruction threshold $\eta \in (0, 1)$ (e.g., $\eta = 0.5$) that marks the point where a frequency becomes "actively restored" by the denoiser. We formally define the \textit{information emergence time} $\tau_\omega$ as the first moment during the reverse generation process (i.e., the largest $t$) where the reconstruction weight reaches this threshold:
\begin{equation}
\tau_\omega := \sup \left\{ t \in (0, T] : H(\omega, t) \ge \eta \right\}.
\end{equation}
\textit{Note:} Since the reverse process moves from $T \to 0$, a larger value of $\tau_\omega$ physically means the event occurs \textbf{earlier} in the generation timeline.

At the critical emergence time $t = \tau_\omega$, the condition implies:
\begin{equation}
\frac{\text{SNR}(\omega, \tau_\omega)}{1 + \text{SNR}(\omega, \tau_\omega)} = \eta \implies \text{SNR}(\omega, \tau_\omega) = \frac{\eta}{1 - \eta}.
\end{equation}
Let $\gamma = \frac{\eta}{1 - \eta} > 0$. Substituting $\text{SNR}(\omega, t) = f(t)\mathcal{S}_{\mathbf{x}_0}(\|\omega\|)$ into the equation, the global scaling factor required to trigger the emergence of frequency $\omega$ is:
\begin{equation}
f(\tau_\omega) = \frac{\gamma}{\mathcal{S}_{\mathbf{x}_0}(\|\omega\|)}.
\end{equation}

\textbf{Theorem 2 (Sequential Emergence of Frequencies in Generation).}
For any two spatial frequencies $\omega_1$ and $\omega_2$, if $\|\omega_1\| > \|\omega_2\|$, the low-frequency component $\omega_2$ is reconstructed strictly earlier in the reverse generation process than the high-frequency component $\omega_1$ (i.e., $\tau_{\omega_2} > \tau_{\omega_1}$).

\textbf{Proof.}
As established in Theorem 1, $f(t) = \frac{\alpha(t)}{1 - \alpha(t)}$ is a strictly monotonically decreasing bijection, meaning its inverse $f^{-1}$ is also strictly monotonically decreasing. 
Assuming $\gamma$ is chosen such that the emergence time is well-defined within $(0, T)$, we can explicitly formulate it as:
\begin{equation}
\tau_\omega = f^{-1} \left( \frac{\gamma}{\mathcal{S}_{\mathbf{x}_0}(\|\omega\|)} \right).
\end{equation}

From Lemma 1, $\|\omega_1\| > \|\omega_2\| \implies \mathcal{S}_{\mathbf{x}_0}(\|\omega_1\|) < \mathcal{S}_{\mathbf{x}_0}(\|\omega_2\|)$. Since both PSD values are strictly positive, taking their reciprocals yields:
\begin{equation}
\frac{\gamma}{\mathcal{S}_{\mathbf{x}_0}(\|\omega_1\|)} > \frac{\gamma}{\mathcal{S}_{\mathbf{x}_0}(\|\omega_2\|)}.
\end{equation}

Applying the strictly monotonically decreasing inverse function $f^{-1}$ to both sides reverses the inequality:
\begin{equation}
f^{-1} \left( \frac{\gamma}{\mathcal{S}_{\mathbf{x}_0}(\|\omega_1\|)} \right) < f^{-1} \left( \frac{\gamma}{\mathcal{S}_{\mathbf{x}_0}(\|\omega_2\|)} \right).
\end{equation}

Substituting the explicit formulation of the emergence time, we obtain:
\begin{equation}
\tau_{\omega_1} < \tau_{\omega_2}.
\end{equation}

\textbf{Physical Implication for Concept Unlearning.}
Because the reverse generation process executes backwards from $t=T$ to $t=0$, the condition $\tau_{\omega_2} > \tau_{\omega_1}$ mathematically guarantees that the network encounters and reconstructs the low-frequency global structure ($\omega_2$) \textbf{strictly earlier} in the generation pipeline than the high-frequency semantic details ($\omega_1$). 
Consequently, forcing unlearning parameter optimization at early generative timesteps (large $t$) forces the model to heavily distort the already-active low-frequency foundational bases ($\Omega_{\text{low}}$) to satisfy the unlearning objective, because the target high-frequency semantics ($\Omega_{\text{semantic}}$) are still suppressed by the Wiener filter ($H \approx 0$) and mathematically inaccessible. This rigorously proves the necessity of placing the Key Step active region strictly at later generative timesteps. 

\begin{table}[t]
\centering
\caption{Quantitative evaluation of Concept Unlearning scaling from a single target to an extreme 50-concept load. We report Unlearn Accuracy (UA $\uparrow$) and Retain Accuracy (RA $\uparrow$). The bold font indicates the best performance, and the underlined font indicates the second-best.}
\setlength{\tabcolsep}{6pt}
\begin{tabular}{l|cc|cc}
\hline
\multirow{2}{*}{\textbf{Method}} & \multicolumn{2}{c|}{\textbf{1 Concept}} & \multicolumn{2}{c}{\textbf{50 Concepts}} \\
\cline{2-5}
 & \textbf{UA(\%)$\uparrow$} & \textbf{RA(\%)$\uparrow$} & \textbf{UA(\%)$\uparrow$} & \textbf{RA(\%)$\uparrow$} \\
\hline
ESD   & \textbf{100.0} & 47.28 & \underline{99.92} & 0.36 \\
GLoCE & 12.14 & \textbf{89.52} & 8.40  & \textbf{90.08} \\
HiRM  & \textbf{100.0} & \underline{88.48} & 23.20 & \underline{84.24} \\
AdaVD & 82.40 & 0.00  & 96.24 & 0.04 \\
DUO   & 37.80 & 77.80 & 98.00 & 0.50 \\
RECE  & \textbf{100.0} & 52.36 & \textbf{100.0} & 1.20 \\
ANT   & {99.10} & 41.20 & 98.50 & 0.20 \\
\hdashline
KSCU  & \textbf{100.0} & 81.12 & 99.52 & 26.40 \\
\hline
\end{tabular}
\label{tab:mass_concept}
\end{table}

\section{More Experimental Details}

The models used in our experiments primarily include Stable Diffusion v1-5 and Stable Diffusion v1-4. KSCU applies targeted fine-tuning based on different unlearning tasks:
\begin{itemize}
    \item \textbf{Class unlearning}: Fine-tuning is conducted on the last 70\% of steps with 700 iterations and a batch size of 1.
    \item \textbf{Style unlearning}: Fine-tuning is applied to the last 50\% of steps with 500 iterations and a batch size of 1.
    \item \textbf{Instance unlearning}: Only the last 20\% of denoising steps are fine-tuned, using 200 iterations.
    \item \textbf{NSFW (nudity) unlearning}: Fine-tuning is applied to the last 70\% of denoising steps with 750 iterations. Unlike other tasks, this process updates all model components except the cross-attention module.
\end{itemize}

All class, style, and instance unlearning tasks involve fine-tuning the model's cross-attention layers, whereas the NSFW unlearning task extends fine-tuning to all components except cross-attention.

For experiments on {Unlearn Canvas}, we use the official implementation provided by {Unlearn Canvas} for all methods except KSCU and ESD. To simplify the experiments while ensuring their validity, we selected 10 classes and 10 styles for evaluation.  
We carefully ensured diversity and representativeness in the chosen classes and styles.  
For experiments on Unlearn Canvas, we followed the original paper and used Stable Diffusion version 1.5.  
However, unlike the original approach, we computed FID using images generated by the frozen model instead of real data. For class and style unlearning, we train 10 models per method, each generating images based on predefined prompts, resulting in a total of $10 \times 51 \times 20 = 10,200$ images. When computing FID, we exclude images corresponding to the 10 target categories entirely and compare the remaining images with those generated by a frozen model.

For experiments on {I2P}, we employed Stable Diffusion version 1.4. The primary reason for this choice is that, compared to later versions, SDv1.4 generates more "nudity" images from I2P prompts, making it more suitable for evaluating the effectiveness of different unlearning methods.  

\textcolor{black}{For instance concept unlearning, we present four comparative cases: two involving human subjects and two involving objects. Meanwhile, utilizing the established celebrity dataset introduced by MACE, we quantitatively compare the results of various methods on both single-instance erasure and 50-instance erasure.}

\section{KSCU Generalization Support Experiment}
\label{sec:appendix_supplementary}

\begin{table}[h]
\color{black}
\centering
\caption{Quantitative performance comparison on the UnlearnCanvas dataset. We report the Unlearn Accuracy (UA), In-domain Retain Accuracy (IRA), and Cross-domain Retain Accuracy (CRA) across three evaluation settings: Original (the initial subset of 10 classes and 10 styles), Supplement (all remaining unexamined concepts in the dataset), and Full (the aggregated performance of the entire benchmark).}
\label{tab:unlearncanvas_additional}
\setlength{\tabcolsep}{4.5pt}
\renewcommand{\arraystretch}{1.15}
\begin{tabular}{c|l|ccc|ccc}
\toprule
\multirow{2}{*}{Setting} 
& \multirow{2}{*}{Method} 
& \multicolumn{3}{c|}{Class} 
& \multicolumn{3}{c}{Style} \\ 
\cmidrule(lr){3-5}\cmidrule(lr){6-8}
& & UA$\uparrow$ & IRA$\uparrow$ & CRA$\uparrow$
& UA$\uparrow$ & IRA$\uparrow$ & CRA$\uparrow$ \\ 
\midrule

\multirow{3}{*}{Original}
& ESD  & 68.6 & 98.9 & 96.4 & 100.0 & 85.7 & 99.0 \\ 
& HiRM & 93.7 & 96.7 & 79.7 & 100.0 & 81.2 & 97.3 \\ 
& KSCU & 70.0 & 99.0 & 97.1 & 100.0 & 88.7 & 99.3 \\ 

\midrule

\multirow{3}{*}{Supplement}
& ESD  & 85.9 & 95.9 & 96.3 & 95.8 & 87.1 & 97.7 \\ 
& HiRM & 98.6 & 72.1 & 64.5  & 99.6 & 65.1 & 72.7 \\ 
& KSCU & 95.6 & 97.5 & 97.1 & 97.0 & 94.3 & 98.6 \\ 

\midrule

\multirow{3}{*}{Full}
& ESD  & 77.3 & 97.4 & 96.3 & 96.6 & 86.9 & 98.0 \\ 
& HiRM & 96.2 & 84.4 & 72.1 & 99.7 & 68.3 & 77.6 \\ 
& KSCU & 82.8 & 98.2 & 97.1 & 97.6 & 93.2 & 98.8 \\ 

\bottomrule
\end{tabular}
\end{table}

\textcolor{black}{\textbf{Validation of Subset Representativeness.} To validate the representativeness of our initial UnlearnCanvas subset, we evaluated all remaining unexamined concepts (``Supplement''), achieving a complete benchmark evaluation (``Full''). Table \ref{tab:unlearncanvas_additional} compares KSCU against ESD and HiRM.}
\textcolor{black}{As shown in Table \ref{tab:unlearncanvas_additional}, KSCU performs consistently on unseen concepts. While HiRM achieves high UA on the supplement, it suffers catastrophic structural collapse in preservation metrics. In contrast, KSCU maintains competitive UA and significantly outperforms baselines in IRA/CRA across the entire benchmark, validating our subset's representativeness and KSCU's generalizability.}

\begin{table}[t]
\color{black} %
\centering
\caption{FLUX nudity erasure results evaluated by NudeNet on I2P prompts and COCO6k FID. For FLUX.1-dev, KSCU and ESD fine-tuned the attention projection layers in the diffusion transformer, including the two-stream module and the single-stream module.}
\label{tab:flux_nudity}
\setlength{\tabcolsep}{3.2pt}
\renewcommand{\arraystretch}{1.12}
\scriptsize
\begin{tabular}{l|cc|cccc|c}
\toprule
\multirow{2}{*}{Method}
& \multicolumn{2}{c|}{Overall Nudity Detection}
& \multicolumn{4}{c|}{Detected Nude Body Parts}
& \multirow{2}{*}{FID$\downarrow$} \\
\cmidrule(lr){2-3}\cmidrule(lr){4-7}
& Unsafe Img.$\downarrow$ & Regions$\downarrow$
& Breast & Gen. & Buttocks & Anus
& \\
\midrule
Original
& 183 & 332
& 282 & 13
& 37 & 0
& -- \\

ESD
& 95 & 129
& 108 & 3
& 18 & 0
& 63.2 \\

HiRM
& 120 & 195
& 156 & 7
& 32 & 0
& 5.8 \\

KSCU
& 13 & 13
& 6 & 7
& 0 & 0
& 26.4 \\
\bottomrule
\end{tabular}
\end{table}
\textcolor{black}{\textbf{Generalizability to Structurally Distinct Backbones.} To verify that KSCU is not limited to UNet-based architectures, we evaluated the DiT-based FLUX.1 model. As shown in Table \ref{tab:flux_nudity}, KSCU seamlessly adapts to this distinct backbone, reducing unsafe images from 183 to 13. It vastly outperforms ESD and HiRM in unlearning thoroughness while maintaining a strong FID, confirming that KSCU's step scheduling is architecture-agnostic.}

\begin{table}[t]
\color{black}
\centering
\caption{Cross-dataset generalization results of KSCU on artist-style concepts. 
For UnlearnCanvas, we report UA, IRA, and CRA. For ESD Artist Prompts, we report manually reviewed UA results using a 50\% threshold.We selected all artist styles included on both datasets for our experiments.}
\label{tab:cross_dataset_artist}
\setlength{\tabcolsep}{4.5pt}
\renewcommand{\arraystretch}{1.12}
\begin{tabular}{l|ccc|cc}
\toprule
\multirow{2}{*}{Concept}
& \multicolumn{3}{c|}{UnlearnCanvas}
& \multicolumn{1}{c}{ESD Artist Prompts} \\
\cmidrule(lr){2-4}\cmidrule(lr){5-6}
& UA$\uparrow$ & IRA$\uparrow$ & CRA$\uparrow$
& UA$\uparrow$ \\
\midrule
Picasso  & 100.0 & 94.9 & 99.3  &100.0\\
Van Gogh & 100.0 & 92.9 & 99.1  &100.0\\
\midrule
Average  & 100.0 & 93.9 & 99.2&100.0\\
\bottomrule
\end{tabular}
\end{table}
\textcolor{black}{\textbf{Cross-Dataset Generalizability of Active Regions.} To further investigate the generalizability of our identified active regions, we evaluate whether the optimal starting threshold $S$ is dataset-specific or concept-specific. According to our theoretical analysis, the frequency emergence dynamics are inherent to the target visual concept itself, rather than the downstream evaluation dataset. Consequently, concepts sharing the same frequency distribution should adhere to identical emergence thresholds across different datasets.}

\textcolor{black}{To empirically validate this, we conducted a cross-dataset evaluation targeting artist styles (Picasso and Van Gogh) using both the UnlearnCanvas dataset and the independent artist prompts collected by ESD. As presented in Table \ref{tab:cross_dataset_artist}, by applying the exact same 50\% starting threshold ($S$) originally determined for art styles, KSCU consistently achieved a 100\% Unlearn Accuracy (UA) across both distinct datasets. This robust performance decisively demonstrates that ourthe threshold exhibits strong concept-specific generalizability, proving that our mathematically derived active regions seamlessly transfer across datasets without the need for dataset-specific hyperparameter retuning.}
\end{document}